\documentclass[runningheads]{llncs}

\usepackage{eccv}

\usepackage{eccvabbrv}
\usepackage[ruled,vlined,linesnumbered]{algorithm2e}

\usepackage{graphicx}
\usepackage{booktabs}
\usepackage{multirow}
\usepackage{xcolor}
\usepackage[accsupp]{axessibility}  

\usepackage{hyperref}

\usepackage{orcidlink}

\begin{document}

\title{Dreaming the Unseen: World Model-regularized Diffusion Policy for Out-of-Distribution Robustness}

\titlerunning{Dream Diffusion Policy}
\author{Ziou Hu\inst{1}\thanks{Equal contribution} \and
Xiangtong Yao\inst{1}\textsuperscript{$\star$} \and
Yuan Meng\inst{1} \and \\
Zhenshan Bing\inst{2,1} \and
Alois Knoll\inst{1}}

\authorrunning{Ziou Hu, Xiangtong Yao et al.}

\institute{School of Computation, Information and Technology, Technical University of Munich, Garching, Germany \and
State Key Laboratory for Novel Software Technology and the School of Science and Technology, Nanjing University (Suzhou Campus), China}

\maketitle

\begin{abstract}
    Diffusion policies excel at visuomotor control but often fail catastrophically under severe out-of-distribution (OOD) disturbances, such as unexpected object displacements or visual corruptions. To address this vulnerability, we introduce the \textbf{Dream Diffusion Policy (DDP)}, a framework that deeply integrates a diffusion world model into the policy's training objective via a shared 3D visual encoder. This co-optimization endows the policy with robust state-prediction capabilities. When encountering sudden OOD anomalies during inference, DDP detects the real-imagination discrepancy and actively abandons the corrupted visual stream. Instead, it relies on its internal ``imagination''—autoregressively forecasted latent dynamics—to safely bypass the disruption, generating imagined trajectories before smoothly realigning with physical reality. Extensive evaluations demonstrate DDP's exceptional resilience. Notably, DDP achieves a 73.8\% OOD success rate on MetaWorld (vs. 23.9\% without predictive imagination) and an 83.3\% success rate under severe real-world spatial shifts (vs. 3.3\% without predictive imagination). Furthermore, as a stress test, DDP maintains a 76.7\% real-world success rate even when relying entirely on open-loop imagination post-initialization.
\end{abstract}

\section{Introduction}
\label{sec:intro}
Robotic manipulation in unstructured environments demands policies that can transform rich visual observations into precise, temporally extended motor commands. Recent visuomotor methods, like diffusion-based policies, have emerged as a compelling paradigm for this setting: instead of regressing a single action, they model action generation as a conditional denoising process~\cite{ho2020denoising, chi2025diffusion, Ze2024DP3} or via flow matching~\cite{zhang2025flowpolicy}. These policies have demonstrated remarkable dexterity and robustness in intricate, high-dimensional robot manipulation tasks ~\cite{chi2025diffusion, he2025demystifying,wu2025device,tian2025pdfactor}.

However, standard visuomotor imitation learning remains fundamentally brittle due to its strict reliance on continuous, uncorrupted visual streams. In real-world deployments, dynamic disturbances (e.g., severe object displacements or camera occlusions) instantly push observations into out-of-distribution (OOD) states, and without explicit physical reasoning, standard policies cannot recover from this massive covariate shift, inevitably leading to compounding errors and catastrophic failure~\cite{gao2025out, wang2026palm, zhu2023viola}. Existing methods attempt to mitigate these shifts through various strategies, including domain randomization~\cite{tobin2017domain}, test-time adaptation~\cite{wang2020tent,yao2025pick,yao2025inference}, explicit expansion of training datasets with synthetic spatial perturbations~\cite{zhang2024diffusion}, or the application of large pretrained visual models~\cite{du2025dynaguide}. Other approaches enhance representational and physical priors~\cite{ada2024diffusion, huang2025improving, wang2025hierarchical, wu2025afforddp}, or learn explicit inverse recovery policies~\cite{sun2025latent, gao2025out}. While somewhat effective, these mitigations typically introduce highly complex architectural pipelines and aggressively alter expert-derived actions during adaptation. This fundamentally compromises the core objective of imitation learning, as the high-fidelity skills carefully extracted from experts are ultimately overwritten rather than preserved.

To overcome these limitations, we rethink how agents handle corrupted visual states by drawing inspiration from human motor mastery. When environmental visual cues become chaotic or deceptive, the optimal human strategy is often to effectively \textbf{close one's eyes} and rely on internal kinesthetic intuition, guided by a single, unwavering focal point. We translate this philosophy into robotic continuous control, arguing that a diffusion policy can be made substantially more resilient by endowing it with a built-in predictive capability. 

The natural technical counterpart to this internal intuition is a world model. Traditionally, these models predict future latent states to support planning~\cite{ha2018world}, enable highly sample-efficient learning~\cite{seo2023masked}, and equip agents with crucial predictive reasoning~\cite{hafner2023mastering}. Diffusion-based world models have emerged as a powerful tool to flexibly forecast complex physical dynamics~\cite{alonso2024diffusion, chandra2025diwa}, making them ideal candidates to provide predictive intuition. Yet, current methods typically treat them as weakly coupled auxiliary modules. Lacking a mechanism to dynamically substitute corrupted vision with imagined latents, these policies remain fatally tethered to deceptive observations and collapse under severe OOD shifts. To sustain control, the world model must transcend passive regularization and become a tightly integrated, active predictive intuition.

We propose the \textbf{Dream Diffusion Policy (DDP)} to tackle severe out-of-distribution (OOD) disturbances. Rather than passively relying on vulnerable visual streams, DDP tightly couples a diffusion policy~\cite{chi2025diffusion} with a predictive world model~\cite{ding2024diffusion} via a \emph{shared 3D visual encoder}. As the policy learns standard behavior cloning~\cite{chi2025diffusion}, the world model simultaneously predicts future observation latents. This shared representation captures robust geometric priors, empowering the world model to actively assist when physical observations become unreliable. During inference, DDP dynamically shifts between physical reality and internal ``imagination.'' By tracking targets and executing core actions entirely within its hallucinated latent space, the robot safely bypasses corrupted vision and physically recovers the environment to an In-Distribution (ID) state.

We extensively validate DDP across 10 MetaWorld~\cite{yu2020meta} tasks, 3 complex Adroit~\cite{rajeswaran2017learning} tasks, and 3 real-world Franka Panda tasks. While Adroit evaluations confirm our world model's predictive stability in high-DoF continuous control, MetaWorld and real-robot experiments demonstrate DDP's exceptional resilience to severe OOD disruptions—such as sudden displacements and complete visual occlusions. Notably, DDP achieves a $73.8\%$ OOD success rate on MetaWorld (vs. $23.9\%$ for a tracking-augmented baseline) and an $83.3\%$ success rate under severe real-world spatial shifts (vs. $3.3\%$). Furthermore, as a stress test, DDP maintains a remarkable $76.7\%$ real-world success rate even when relying entirely on open-loop imagination post-initialization. In summary, our primary contributions are:
\begin{itemize}
    \item \textbf{World-Model-Regularized Training Pipeline:} We propose the Dream Diffusion Policy, which co-optimizes behavior cloning and world-model objectives via a shared 3D encoder. This unified representation allows the world model's future predictions to directly regularize and guide policy inference.
    \item \textbf{Robust Inference via Predictive Imagination:} We design a novel, subtask-aware execution mechanism that actively detects OOD states via a highly intuitive real-imagination discrepancy ($\mathcal{D}_{R-I}$), bypasses corrupted vision using autoregressive imagination, and triggers a localized physical recovery to the ID state to resume normal execution.
    \item \textbf{State-of-the-Art OOD Robustness:} We demonstrate that DDP significantly outperforms state-of-the-art baselines under severe OOD conditions across both simulated and real-world tasks, maintaining exceptional resilience even when operating completely blind after initial observation.
\end{itemize}

\section{Related Works}
\label{sec:related_works}

\noindent \textbf{Diffusion Policy for Robotic Manipulation.} By framing action generation as conditional denoising, diffusion models~\cite{ho2020denoising} effectively capture complex, multimodal distributions in high-dimensional continuous control~\cite{chi2025diffusion,Ze2024DP3}. These policies achieve state-of-the-art results across diverse applications, including kinematics-aware~\cite{ma2024hierarchical}, contact-rich~\cite{10912754,xue2025reactive}, and cross-manipulator tasks~\cite{yao2025inference, 10855557}. Additionally, mechanisms like receding-horizon control and stochastic Langevin dynamics provide inherent robustness against minor physical perturbations and transient visual distractions in real-world deployments~\cite{chi2025diffusion,xue2025reactive}.

\noindent \textbf{World Models for Decision Making and Robotics.} 
World models learn internal representations to predict environmental dynamics and guide decision-making~\cite{ding2025understanding}. Pioneered by the Dreamer series~\cite{hafner2019dream, hafner2020mastering, hafner2023mastering}, agents can master complex continuous control entirely within a learned latent imagination. This paradigm is now advancing embodied AI: models like NVIDIA Cosmos~\cite{agarwal2025cosmos} enforce strict physical laws as ``physical AI'' brains, while architectures like DreamZero~\cite{ye2026world} jointly forecast future states and motor actions. Ultimately, integrating predictive world models directly into control policies provides the crucial ``imagination'' required for robust robotic manipulation.

\noindent \textbf{OOD Challenges and Mitigation Strategies.} Deployed policies inevitably face Out-of-Distribution (OOD) shifts. Standard mitigations expand training diversity via domain randomization~\cite{tobin2017domain}, synthetic trajectories~\cite{huang2025improving}, or reinforcement learning (RL)~\cite{wang2025hierarchical}. However, these methods fail under severe conditions—like camera occlusions or target shifts—because they assume a continuous visual stream. Without visual evidence, policies suffer from shortcut learning~\cite{geirhos2020shortcut}, fatally over-relying on proprioception for spatial localization. Furthermore, RL fine-tuning~\cite{curtis2025flowbased,chandra2025diwa} cannot reconstruct missing visual data at inference, while heavy online pipelines~\cite{du2025dynaguide} and video-diffusion reward models~\cite{huang2024diffusion} are computationally prohibitive. 

While world models~\cite{ha2018world} and diffusion variants~\cite{ding2024diffusion} forecast states to handle partial observability, current integrations treat them as weakly coupled auxiliary modules that still collapse under severe shifts. To overcome this, we embed lightweight predictive capabilities directly into the policy. By co-optimizing a diffusion policy and world model via a shared-encoder regularizer, we avoid complex decoupled architectures and expensive RL fine-tuning~\cite{chandra2025diwa}. This tight coupling empowers the policy to seamlessly ``dream'' missing visual context, sustaining robust control even during severe sensor dropouts.

\section{Methods}

\begin{figure}[tb]
  \centering
  \includegraphics[height=5.7cm]{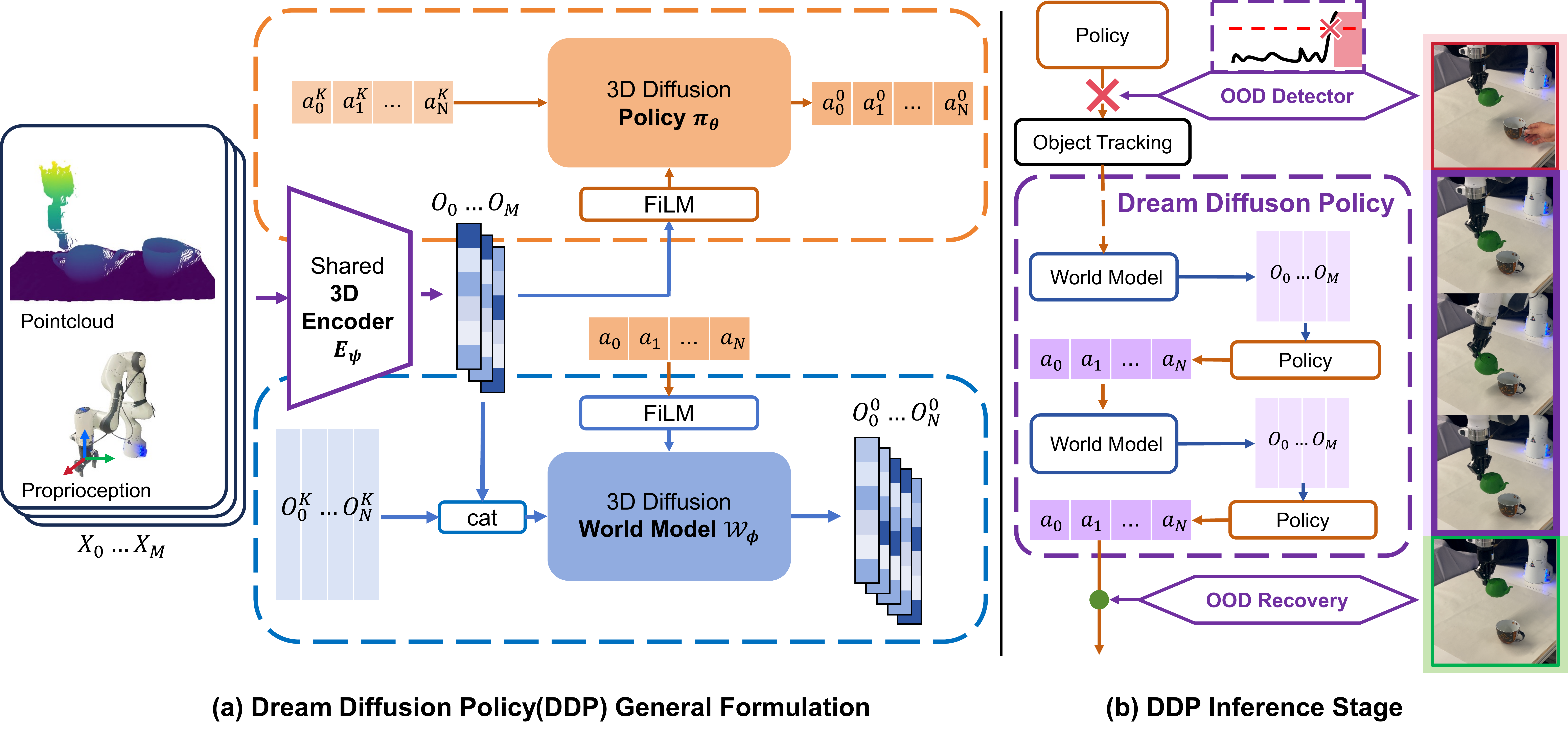}
  \caption{\textbf{Overview of the Dream Diffusion Policy (DDP) framework.} \textbf{(a)} A shared 3D encoder extracts unified latent embeddings from point clouds and proprioception. These history latents jointly condition the Diffusion Policy to denoise action chunks, and the Diffusion World Model (via channel-wise concatenation) to forecast future latent states. \textbf{(b)} During inference stage, upon detecting an Out-of-Distribution anomaly, the system halts, tracks the object, and transitions into an imagination loop. The World Model autoregressively forecasts latent states to drive the policy's action generation, safely bypassing corrupted observations until the recovery is triggered.}
  \label{fig:overview}
\end{figure}

Building on DP3 \cite{Ze2024DP3}, our Dream Diffusion Policy (DDP) utilizes a shared 3D encoder to extract unified geometric latent embeddings from point clouds and robot proprioception. These embeddings jointly condition both the Diffusion Policy and the Diffusion World Model. During training, the World Model loss acts as a regularizer for the policy, embedding environmental kinematics and dynamics into the shared representation. During inference, this coupled architecture empowers the agent to use ``Imagination''---forecasting future observations---to sustain robust execution under Out-of-Distribution scenarios, as outlined in \cref{fig:overview}.

\begin{algorithm}[t]
\caption{Dream Diffusion Policy Training}
\label{alg:dream_diffusion_training}
\KwIn{Dataset $\mathcal{D}$, Max steps $K$, 3D Encoder $E_\psi$, Policy $\pi_\theta$, World Model $\mathcal{W}_\phi$}

\While{not converged}{
    \tcp{1. Sample Trajectories($M$: observation steps, $N$: action steps, $H$: horizon)}
    Sample $(X_{0 \dots M+N-1}, a_{0 \dots H-1}) \sim \mathcal{D}$\;
    
    \tcp{2. Encode Observations}
    $O_{0 \dots M-1} \leftarrow E_\psi(X_{0 \dots M-1})$\;
    $O_{M \dots M+N-1} \leftarrow E_\psi(X_{M \dots M+N-1})$\;
    
    \tcp{3. Diffusion Policy Loss}
    $k_p \sim \mathcal{U}(1, K), \epsilon_p \sim \mathcal{N}(0, \mathbf{I})$\;
    $a^{k_p}_{0 \dots H-1} \leftarrow \text{AddNoise}(a_{0 \dots H-1}, \epsilon_p, k_p)$\;
    $\mathcal{L}_{BC} \leftarrow \text{MSE}\left(\epsilon_p, \pi_\theta(a^{k_p}_{0 \dots H-1}, k_p, \text{FiLM}(O_{0 \dots M-1}))\right)$\;
    
    \tcp{4. Diffusion World Model Loss}
    $k_w \sim \mathcal{U}(1, K), \epsilon_w \sim \mathcal{N}(0, \mathbf{I})$\;
    $O^{k_w}_{M \dots M+N-1} \leftarrow \text{AddNoise}(O_{M \dots M+N-1}, \epsilon_w, k_w)$\;
    $C_{wm} \leftarrow \text{cat}(O_{0 \dots M-1}, O^{k_w}_{M \dots M+N-1})$\;
    $\mathcal{L}_{WM} \leftarrow \text{MSE}\left(\epsilon_w, \mathcal{W}_\phi(C_{wm}, k_w, \text{FiLM}(a_{M-1 \dots M+N-2}))\right)$\;
    
    \tcp{5. Optimization}
    $\mathcal{L}_{total} \leftarrow \mathcal{L}_{BC} + \lambda \mathcal{L}_{WM}$\;
    Update $\psi, \theta, \phi$ using $\nabla_{\psi, \theta, \phi} \mathcal{L}_{total}$\;
}
\end{algorithm}

\label{sec:methods}

\subsection{Action and Latent Embedding Chunking}
Action chunking acts as policy compression in human motor control \cite{lai2022action}. In robotic imitation learning, it better captures temporal dependencies \cite{liu2024bidirectional} and mitigates compounding errors common in strictly autoregressive models \cite{cen2025worldvla}. Motivated by this, we synchronize our policy and world model via explicit chunking:

\textbf{Action Chunking:} Our Diffusion Policy denoises a full trajectory over horizon $H$. To balance long-term consistency and local reactivity, the agent executes only a sub-sequence ``chunk'' of length $N$.

\textbf{Latent Embedding Observation Chunking:} To enable forward-looking ``imagination'', the World Model operates over the same chunk. Given $M$ historical observations and the $N$-step planned action chunk, it predicts $N$ future latent states in parallel. For continuous OOD rollouts, we apply a sliding window: the final $M$ predicted latents become the historical condition for the next cycle, tightly aligning planned actions with expected outcomes.

\subsection{Diffusion Policy}
Building upon standard DDPM \cite{ho2020denoising}, our diffusion policy $\pi_\theta$ formulates action generation as a conditional denoising task, predicting an action sequence $\mathbf{a}_{0 \dots H-1}$ over horizon $H$. Following DP3 \cite{Ze2024DP3}, we utilize a shared 3D encoder $E_\psi$ to process raw historical observations $\mathbf{X}_{0 \dots M-1}$. This encoder extracts spatial features from geometric point clouds via PointNet and encodes proprioceptive states (e.g., end-effector pose) via an MLP, concatenating both streams to yield the unified context $\mathbf{O}_{0 \dots M-1}$. During the forward diffusion process, Gaussian noise $\epsilon_p \sim \mathcal{N}(0, \mathbf{I})$ is added to the ground-truth actions to generate a noisy trajectory $\mathbf{a}^{k_p}$ at a randomly sampled timestep $k_p \sim \mathcal{U}(1, K)$.

The core of the policy is a 1D Conditional U-Net. The encoded history context $\mathbf{O}_{0 \dots M-1}$ serves as a global condition. It is injected into the intermediate residual blocks of the U-Net via Feature-wise Linear Modulation (FiLM) \cite{perez2018film}, allowing the visual state to guide the action denoising process effectively. The objective function for the policy is standard Behavior Cloning (BC) via noise matching:
\begin{equation}
    \mathcal{L}_{BC} = \mathbb{E}_{\mathbf{a}, \mathbf{X}, \epsilon_p, k_p} \left[ \left\| \epsilon_p - \pi_\theta \left(\mathbf{a}^{k_p}_{0 \dots H-1}, k_p, \text{FiLM}(\mathbf{O}_{0 \dots M-1}) \right) \right\|_2^2 \right]
\end{equation}

\subsection{Diffusion World Model}
Based on the framework in \cite{ding2024diffusion}, our Diffusion World Model $\mathcal{W}_\phi$ predicts $N$-step future latent embeddings $\mathbf{O}_{M \dots M+N-1}$. During training, the shared 3D encoder $E_\psi$ extracts these future targets, which are then corrupted with noise $\epsilon_w$ at timestep $k_w$ to yield $\mathbf{O}^{k_w}_{M \dots M+N-1}$. Because actions (directional controls) and observations (geometric context) possess fundamentally distinct properties, we deviate from the sampling-based ``prefix-fixing'' conditioning used in \cite{ding2024diffusion}. We argue that fixing initial states during reverse diffusion lacks the persistent grounding required for high-dimensional point cloud latents. Instead, we employ direct \textbf{feature-level integration} \cite{saharia2022palette}. By repeating the historical latent $\mathbf{x}_{obs}$ across the temporal horizon and concatenating it channel-wise with the noisy samples, we formulate the input as:
\begin{equation}
    \mathbf{X}_{in} = \text{Repeat}(\mathbf{x}_{obs}, N) \oplus \mathbf{X}_{sample}
\end{equation}
This ensures every convolutional layer in the 1D U-Net has immediate, continuous access to the historical structural context.

Concurrently, the executed action sequence $\mathbf{a}_{M-1 \dots M+N-2}$ serves as a global condition. As in the policy architecture, these action representations are fused with the diffusion timestep embeddings and injected via FiLM. The world model is optimized by minimizing the noise prediction error:
\begin{equation}
    \mathcal{L}_{WM} = \mathbb{E}_{\mathbf{O}, \mathbf{a}, \epsilon_w, k_w} \left[ \left\| \epsilon_w - \mathcal{W}_\phi \left( \mathbf{C}_{wm}, k_w, \text{FiLM}(\mathbf{a}_{M-1 \dots M+N-2}) \right) \right\|_2^2 \right]
\end{equation}

\subsection{Training Objective}
We jointly optimize the Diffusion Policy and World Model end-to-end. Using normalized expert demonstrations, the shared encoder $E_\psi$ extracts latent embeddings for both historical context and future targets. The policy $\pi_\theta$ learns to predict Gaussian noise added to the action horizon $\mathbf{a}_{0 \dots H-1}$ conditioned on the history $\mathbf{O}_{0 \dots M-1}$. Simultaneously, the world model $\mathcal{W}_\phi$ predicts the noise injected into future observation latents $\mathbf{O}_{M \dots M+N-1}$, conditioned on the history and executed actions. Following established regularization strategies \cite{huang2025improving, ada2024diffusion}, the world model acts as a geometric and dynamic regularizer. We update the network parameters $(\psi, \theta, \phi)$ via gradient descent using a combined loss:
\begin{equation}
    \mathcal{L}_{total} = \mathcal{L}_{BC} + \lambda \mathcal{L}_{WM}
\end{equation}
where $\lambda$ balances the two objectives. Finally, we apply an Exponential Moving Average (EMA) \cite{morales2024exponential} to the model weights to stabilize convergence and enhance inference robustness (\cref{alg:dream_diffusion_training}).

\subsection{Inference Stage}

During inference, dynamic object displacements break the In-Distribution (ID) environment, triggering an Out-of-Distribution (OOD) state. To maintain spatial awareness during these disruptions, we integrate a 6D pose estimator (e.g., FoundationPose \cite{wen2024foundationpose}). By combining this estimator with the World Model's predictive ``imagination,'' we introduce a hybrid mechanism of detection, recursive imagination, and physical recovery to safely bridge OOD gaps, as illustrated in the closed-loop workflow of \cref{fig:inference}. The pseudocode for closed-Loop inference is provided in \cref{alg:inference_stage} in Appendix.

\noindent \textbf{Subtask Definition}
For long-horizon manipulation, we define a subtask sequence $\mathcal{S} = \{S_1, \dots, S_J\}$, each targeting an object $C_j$. Lacking explicit dataset annotations, we bypass complex clustering and use simple expert heuristics for completion triggers (e.g., gripper closure for picking, allocated time for pouring). This flexibility allows DDP to solve complex, multi-stage challenges.

\noindent \textbf{Real-Imagination Discrepancy as OOD Detector.}
Unlike methods relying on reconstruction errors \cite{isaku2025out, graham2023denoising} or diffusion loss \cite{lee2025diff}, we take a highly intuitive approach: detecting OOD anomalies via the latent discrepancy between the real observation $\mathbf{O}_{real}^t = E_\psi(X_{real}^t)$ and the World Model's prediction $\mathbf{O}_{pred}^t$:
\begin{equation}
    \mathcal{D}_{R-I}(t) = \| \mathbf{O}_{real}^t - \mathbf{O}_{pred}^t \|_2^2
\end{equation}
Nominal actions maintain a low $\mathcal{D}_{R-I}$. When an external disturbance pushes $\mathcal{D}_{R-I}(t)$ above a threshold $\tau_{diff}$, an OOD event is flagged. To identify \textit{which} objects moved, we query the pose estimator, comparing current poses to expected ID states to isolate the displaced objects $\mathcal{C}_{disp}$.

\noindent \textbf{Simplified Object Tracking under OOD.}
Rather than relying on continuous tracking during chaotic displacements, we assume the displaced object $C_j$ eventually settles into a static pose. Once motion ceases, we query the 6D pose estimator for its new resting pose $\mathbf{p}_{C_j} \in SE(3)$. By comparing this to its expected ID pose $\mathbf{p}_{C_j}^{exp}$, we compute a single static spatial displacement:
\begin{equation}
    \Delta \mathbf{p}_{C_j} = \mathbf{p}_{C_j} - \mathbf{p}_{C_j}^{exp}
\end{equation}
To re-align the robot with the displaced target, we directly apply this offset as a compensatory tracking action for the end-effector, defined as $\mathbf{a}^{track}_j = \Delta \mathbf{p}_{C_j}$. 

\noindent \textbf{Recursive Imagination for Action Generation}
To bypass corrupted physical observations during an OOD state, we implement a recursive ``wheel'' mechanism for imagined execution after tracking (\cref{fig:overview}(b)). Instead of using real visual inputs, the policy generates action chunks ($a_0 \dots a_N$) conditioned entirely on the previously predicted latent embedding. Concurrently, the Diffusion World Model uses this imagined latent and the newly predicted actions to forecast the \textit{next} latent state ($O_0 \dots O_M$). Autoregressively feeding these forecasts back into the policy sustains a coherent trajectory purely within the agent's internal imagination, enabling safe task execution until target recovery is triggered.

\noindent \textbf{Target-Specific Recovery.}
Since simultaneous global recovery is impractical, we restore the environment iteratively. For an imagined subtask $S_j$ targeting $C_j \in \mathcal{C}_{disp}$, we define a simple heuristic boolean trigger $\mathcal{T}_{end}(S_j)$. Once $\mathcal{T}_{end}(S_j) = \text{True}$, the agent applies an inverse spatial translation $\mathbf{a}^{rec}_j = -\Delta \mathbf{p}_{C_j}$ to realign the imagined trajectory with the physical object, removing $C_j$ from $\mathcal{C}_{disp}$. We then re-evaluate $\mathcal{D}_{R-I}$: if it remains above $\tau_{diff}$, the policy continues executing via imagination. Otherwise, the environment is fully recovered, and the policy reverts to the real visual latent $\mathbf{O}_{real}^t$ for standard ID execution.

\begin{figure}[tb]
  \centering
  \includegraphics[height=3.5cm]{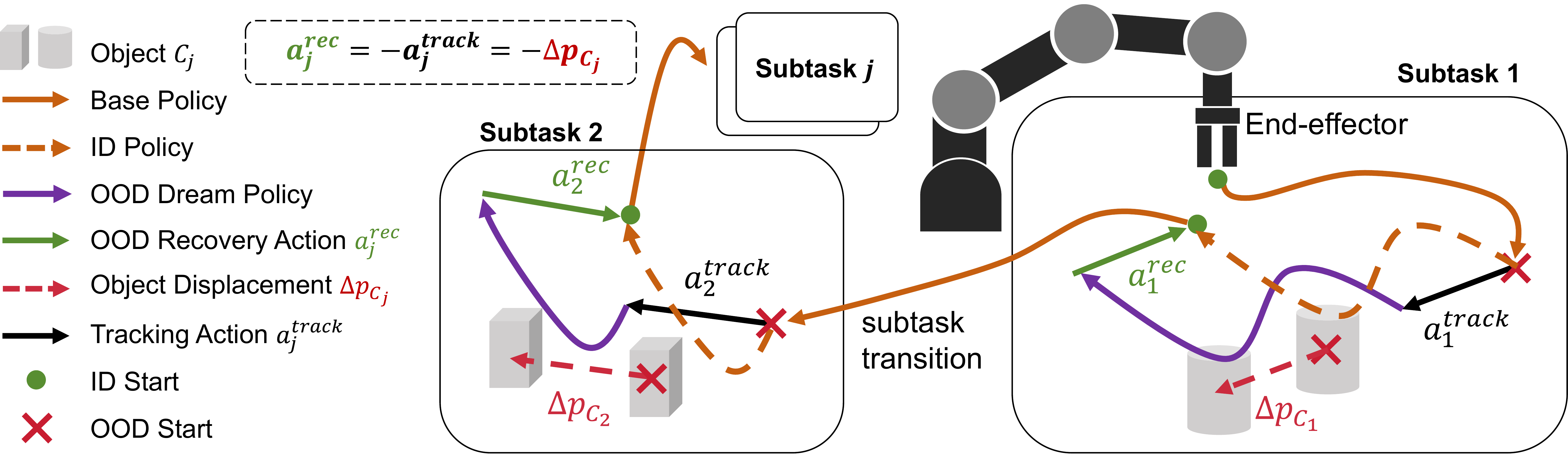}
  \caption{\textbf{Overview of the DDP workflow.} Illustrating the closed-loop transitions between physical reality (ID) and internal imagination (OOD) across subtasks.}
  \label{fig:inference}
\end{figure}

\section{Experiments}
\label{sec:experiment}

\subsection{Simulation Experiments}

\noindent \textbf{Benchmarks and Datasets}
To evaluate the proposed method, we utilize two widely adopted continuous control simulation benchmarks: MetaWorld \cite{yu2020meta} and Adroit \cite{rajeswaran2017learning}. Specifically, we construct our evaluation pool using 10 MetaWorld tasks, ranging from simple to hard difficulties, to test general robotic manipulation capabilities. Additionally, to assess performance on highly complex dynamical systems, we select 3 tasks from the Adroit benchmark.

\noindent \textbf{Model Setup}
The shared 3D visual encoder processes $M=2$ observation steps, concatenating 64-dimensional features from both a PointNet (point cloud) and an MLP (proprioception) into a 128-dimensional unified latent embedding. Conditioned on this $M=2$ history, the Diffusion Policy predicts an action horizon of $H=16$ and executes an $N=8$ step chunk. It employs a 1D U-Net with down-sampling channels of $[256, 512, 1024]$ and a 128-dimensional diffusion step embedding. Concurrently, the Diffusion World Model uses a separate, identically parameterized 1D U-Net to predict $N=8$ future latent embeddings conditioned on the $M=2$ history and $N=8$ planned actions. $\lambda$ is set to 1.

\begin{figure}[tb]
  \centering
  \includegraphics[height=5cm]{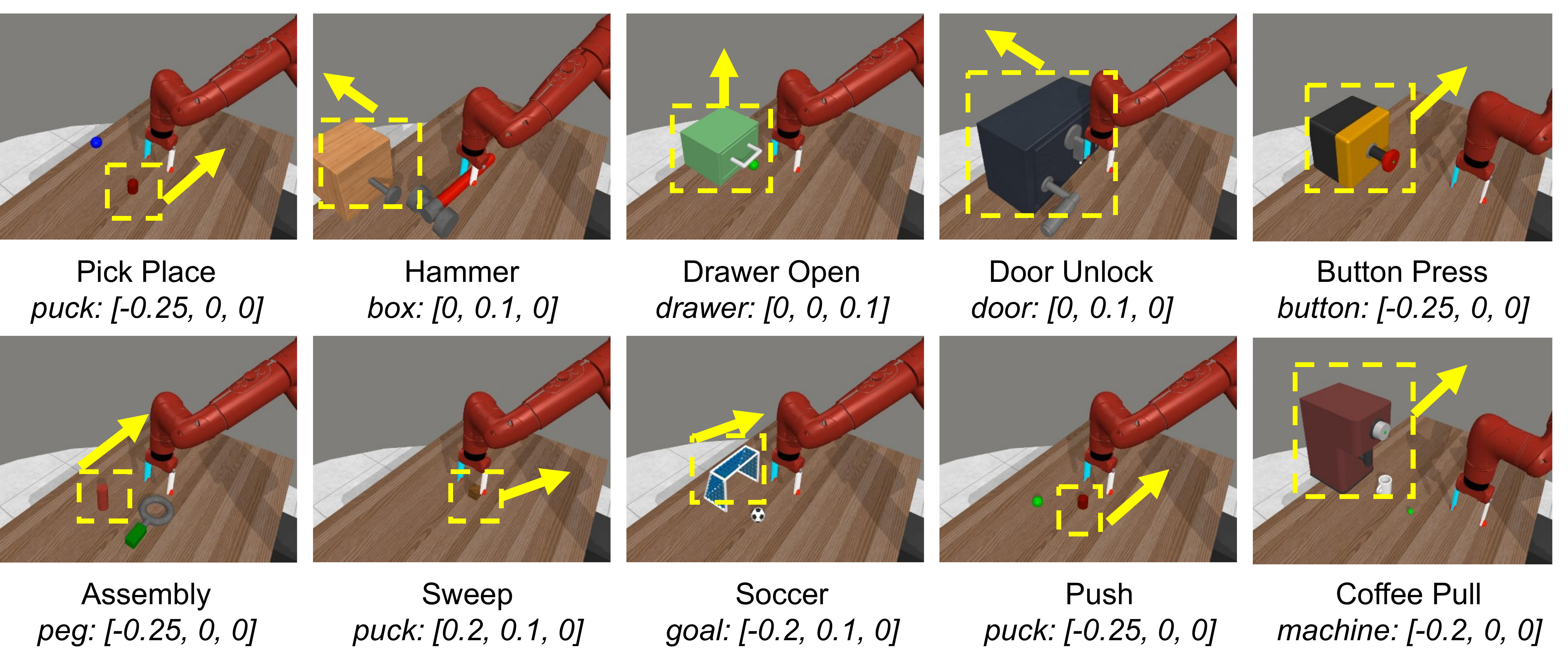}
  \caption{\textbf{Out-of-Distribution (OOD) spatial disturbances in MetaWorld.} \textbf{Yellow dashed boxes} highlight target objects, and \textbf{solid arrows} indicate the direction of dynamic displacements injected mid-task. The exact linear translation vectors $(x, y, z)$ applied to shift each object out-of-distribution are detailed below.}
  \label{fig:ood_disturbances}
\end{figure}

\noindent \textbf{Out-of-Distribution Setup}
To test robustness against spatial shifts, we evaluate OOD performance exclusively in MetaWorld. During training, targets are randomized within a fixed In-Distribution area. For OOD inference, we introduce sudden mid-task displacements, moving objects entirely outside these training boundaries (\cref{fig:ood_disturbances}). Applied as direct linear coordinate translations in MuJoCo, these shifts provide measurable disturbances to reliably evaluate recovery.

\noindent \textbf{Training and Evaluation}
For a fair and rigorous comparison, we benchmark our Dream Diffusion Policy (DDP) against two state-of-the-art baselines: DP3~\cite{Ze2024DP3} and FlowPolicy~\cite{zhang2025flowpolicy}. All models are trained for 500 epochs with 3 seeds (seed0, 1, 2) using a fixed dataset of 50 demonstrations per task. To ensure an equitable evaluation, the foundational policy architectures and core hyperparameters across all methods are kept strictly identical. During inference, we evaluate each method for 100 episodes per task, per seed, and calculate the average success rate. Our evaluation is systematically divided into the following phases:
\begin{enumerate}
    \item \textbf{Base Policy Accuracy (ID \& OOD):} Evaluating standard In-Distribution and Out-of-distribution performance of DP3, FlowPolicy, and DDP.
    \item \textbf{Dream Diffusion Policy Accuracy (OOD):} Evaluating DDP using our tracking, imagination and recovery mechanism under OOD conditions.
    \item \textbf{Ablation Tracking-Augmented Accuracy (OOD):} To isolate the benefits of our World Model, we perform a direct ablation by applying the tracking intervention to the base DP3 policy. We additionally evaluate FlowPolicy equipped with the same tracking mechanism for a broader comparison.
    \item \textbf{World Model Imagination Analysis (ID):} To test the World Model's reliability, we randomly replace real observation with imagined latents after the initial chunk at $10\%$, $50\%$ and $100\%$ (open-loop imagined rollout) probabilities.
\end{enumerate}

\begin{table*}[tb]
  \caption{\textbf{Evaluation Results (Success Rate \%).} We report the In-Distribution (ID) and Out-of-Distribution (OOD) performance of the base policies, the tracking-augmented policies (DP3 ablation, Flow Policy, our DDP) under OOD, and the robustness of DDP under various imagination substitution rates.}
  \label{tab:eval_results}
  \centering
  \resizebox{\textwidth}{!}{
  \begin{tabular}{l | c c c | c c c c c c c c c c | c c c}
    \toprule
    \textbf{Algorithm /}  & \multicolumn{3}{c|}{\bf Adroit} & \multicolumn{10}{c|}{\bf MetaWorld} & \textbf{Adroit} & \textbf{MetaWorld} & \textbf{Total} \\
    \textbf{Condition}  & Door & Hammer & Pen & Assembly & \begin{tabular}{@{}c@{}}Button \\ Press\end{tabular} & \begin{tabular}{@{}c@{}}Coffee \\ Pull\end{tabular} & \begin{tabular}{@{}c@{}}Door \\ Unlock\end{tabular} & \begin{tabular}{@{}c@{}}Drawer \\ Open\end{tabular} & Hammer & \begin{tabular}{@{}c@{}}Pick \\ Place\end{tabular} & Push & Soccer & Sweep & \textbf{Avg} & \textbf{Avg} & \textbf{Avg} \\
    \midrule
    \multicolumn{17}{l}{\textbf{Base Policy (ID)}} \\
    \midrule
    DP3 & $69\pm4$ & $100\pm0$ & $76\pm7$ & $100\pm0$ & $100\pm0$ & $99\pm1$ & $100\pm0$ & $100\pm0$ & $100\pm0$ & $95\pm1$ & $94\pm1$ & $35\pm3$ & $100\pm0$ & 81.5 & 92.3 & 89.8 \\
    FlowPolicy & $64\pm4$ & $88\pm1$ & $67\pm4$ & $100\pm0$ & $100\pm0$ & $100\pm0$ & $100\pm0$ & $100\pm0$ & $98\pm1$ & $96\pm1$ & $95\pm1$ & $53\pm6$ & $100\pm0$ & 72.9 & 94.2 & 89.3 \\
    \textbf{DDP (Ours)} & $63\pm3$ & $99\pm1$ & $71\pm2$ & $100\pm0$ & $100\pm0$ & $100\pm0$ & $100\pm0$ & $100\pm0$ & $100\pm0$ & $95\pm1$ & $93\pm1$ & $42\pm7$ & $100\pm0$ & 77.8 & 93.0 & 89.5 \\
    \midrule
    \multicolumn{17}{l}{\textbf{Base Policy (OOD)}} \\
    \midrule
    DP3 & - & - & - & $0\pm0$ & $0\pm0$ & $0\pm0$ & $2\pm2$ & $4\pm2$ & $0\pm0$ & $0\pm0$ & $0\pm0$ & $1\pm1$ & $0\pm0$ & - & 0.8 & 0.8 \\
    FlowPolicy & - & - & - & $0\pm0$ & $11\pm5$ & $0\pm0$ & $53\pm14$ & $0\pm0$ & $1\pm1$ & $0\pm0$ & $0\pm0$ & $2\pm2$ & $0\pm0$ & - & 6.7 & 6.7 \\
    \textbf{DDP (Ours)} & - & - & - & $0\pm0$ & $0\pm0$ & $0\pm0$ & $5\pm5$ & $3\pm0$ & $1\pm1$ & $0\pm0$ & $0\pm0$ & $4\pm2$ & $0\pm0$ & - & 1.3 & 1.3 \\
    \midrule
    \multicolumn{17}{l}{\textbf{Tracking-Augmented (OOD)}} \\
    \midrule
    FlowPolicy & - & - & - & $0\pm0$ & $13\pm2$ & $0\pm0$ & $41\pm5$ & $0\pm0$ & $0\pm0$ & $0\pm0$ & $0\pm0$ & $0\pm0$ & $0\pm0$ & - & 5.4 & 5.4 \\
    DP3 (Ablation) & - & - & - & $0\pm0$ & $93\pm3$ & $23\pm4$ & $59\pm16$ & $3\pm2$ & $1\pm0$ & $13\pm13$ & $9\pm3$ & $16\pm4$ & $22\pm14$ & - & 23.9 & 23.9 \\
    \textbf{DDP (Ours)} & - & - & - & $96\pm2$ & $100\pm0$ & $94\pm3$ & $74\pm4$ & $100\pm0$ & $100\pm0$ & $39\pm5$ & $39\pm6$ & $17\pm4$ & $79\pm7$ & - & \textbf{73.8} & \textbf{73.8} \\
    \midrule
    \multicolumn{17}{l}{\textbf{DDP Imagination Interventions (ID)}} \\
    \midrule
    \textbf{DDP (10\% Imagined)} & $95\pm3$ & $99\pm0$ & $92\pm1$ & $100\pm0$ & $100\pm0$ & $100\pm0$ & $99\pm0$ & $98\pm1$ & $100\pm0$ & $93\pm3$ & $93\pm2$ & $40\pm3$ & $100\pm0$ & \textbf{95.2} & \textbf{92.3} & \textbf{93.0} \\
    \textbf{DDP (50\% Imagined)} & $96\pm1$ & $98\pm2$ & $87\pm3$ & $99\pm0$ & $100\pm0$ & $98\pm1$ & $99\pm1$ & $90\pm1$ & $100\pm0$ & $81\pm6$ & $91\pm2$ & $39\pm3$ & $98\pm1$ & \textbf{94.0} & \textbf{89.4} & \textbf{90.5} \\
    \textbf{DDP (Open-loop)} & $96\pm3$ & $98\pm2$ & $86\pm3$ & $18\pm2$ & $100\pm0$ & $54\pm3$ & $97\pm1$ & $48\pm3$ & $99\pm0$ & $21\pm3$ & $38\pm3$ & $42\pm3$ & $97\pm1$ & \textbf{93.2} & \textbf{61.4} & \textbf{68.7} \\
    \bottomrule
  \end{tabular}
  }
\end{table*}

\noindent \textbf{Results}
As shown in \cref{tab:eval_results}, standard base policies fail under severe OOD shifts. However, our tracking-augmented DDP effectively handles these disturbances, achieving a 73.8\% average success rate on MetaWorld. Applying identical tracking to the baselines yields lower success rates (23.9\% for DP3, 5.4\% for FlowPolicy). This proves tracking alone is insufficient; the Diffusion World Model's coupled latents are crucial for robust task completion.

\begin{figure}[tb]
  \centering
  \includegraphics[height=2.4cm]{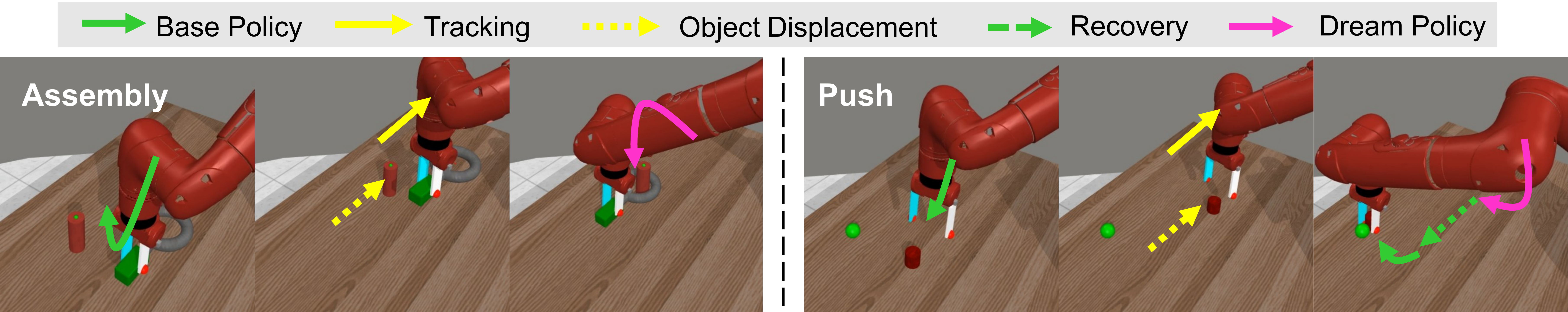}
    \caption{\textbf{DDP execution on MetaWorld Assembly and Push.} Sequences show the transition from the ID Base Policy (green) to OOD tracking (yellow), bypassing visual shifts via the imagined Dream Policy (pink), and finally recovering to the target.}
  \label{fig:qualitative_results}
\end{figure}

\Cref{fig:qualitative_results} illustrates DDP's execution on the MetaWorld Assembly and Push tasks. In Assembly (2 subtasks), the second object peg is displaced Out-of-Distribution after the robot grasps the ring. After tracking this shift, DDP executes a precise action to successfully insert the ring. In Push (2 subtasks), the first object puck is displaced before grasping. DDP tracks the shifted puck, execute actions via imagination, utilizes a localized recovery action to move the end-effector to ID position, and accurately pushes it to the target. For more comprehensive results, please refer to \cref{sec:appendix_metaworld} in the Appendix.

Furthermore, \cref{tab:eval_results} demonstrates DDP's resilience when replacing real observations with imagined latents. On MetaWorld, DDP maintains near-baseline accuracy at 10\% (92.3\%) and 50\% (89.4\%) substitution rates, remaining robust even at 100\% (61.4\%). On the complex Adroit benchmark, imagination actually \textit{boosts} success from a 77.8\% baseline to 95.2\% at 10\% substitution. This confirms the built-in World Model not only bridges visual gaps but also acts as a temporal smoother, stabilizing high-DoF continuous control.

\subsection{Real World Experiments}

\noindent \textbf{Real Robot Benchmark}
To evaluate physical efficacy and performance under OOD, we design three real-world manipulation tasks of varying complexity. During training, initial target positions are uniformly randomized along a 10 cm linear range to ensure baseline state variance.
\begin{itemize}
    \item \textbf{Press Button (Easy | 1 Subtask, 1 OOD Target):} The robot closes its gripper to precisely press a button. Success requires depressing it exactly to the activation threshold, avoiding both incomplete contact (under-pressing) and excessive force (over-pressing).
    
    \item \textbf{Pour Tea (Medium | 2 Subtasks, 1 OOD Target):} The robot grasps a teapot, navigates to a cup, executes a pouring motion, and returns the teapot to the table. Success requires the spout to remain precisely aligned with the cup's center during the entire pour.
    
    \item \textbf{Stack Blocks (Hard | 2 Subtasks, 2 OOD Targets):} The robot picks up a pink block and places it atop a green base block. Success requires the pink block to remain perfectly stable and balanced after the gripper releases.
\end{itemize}

\begin{figure}[tb]
  \centering
  \includegraphics[height=5cm]{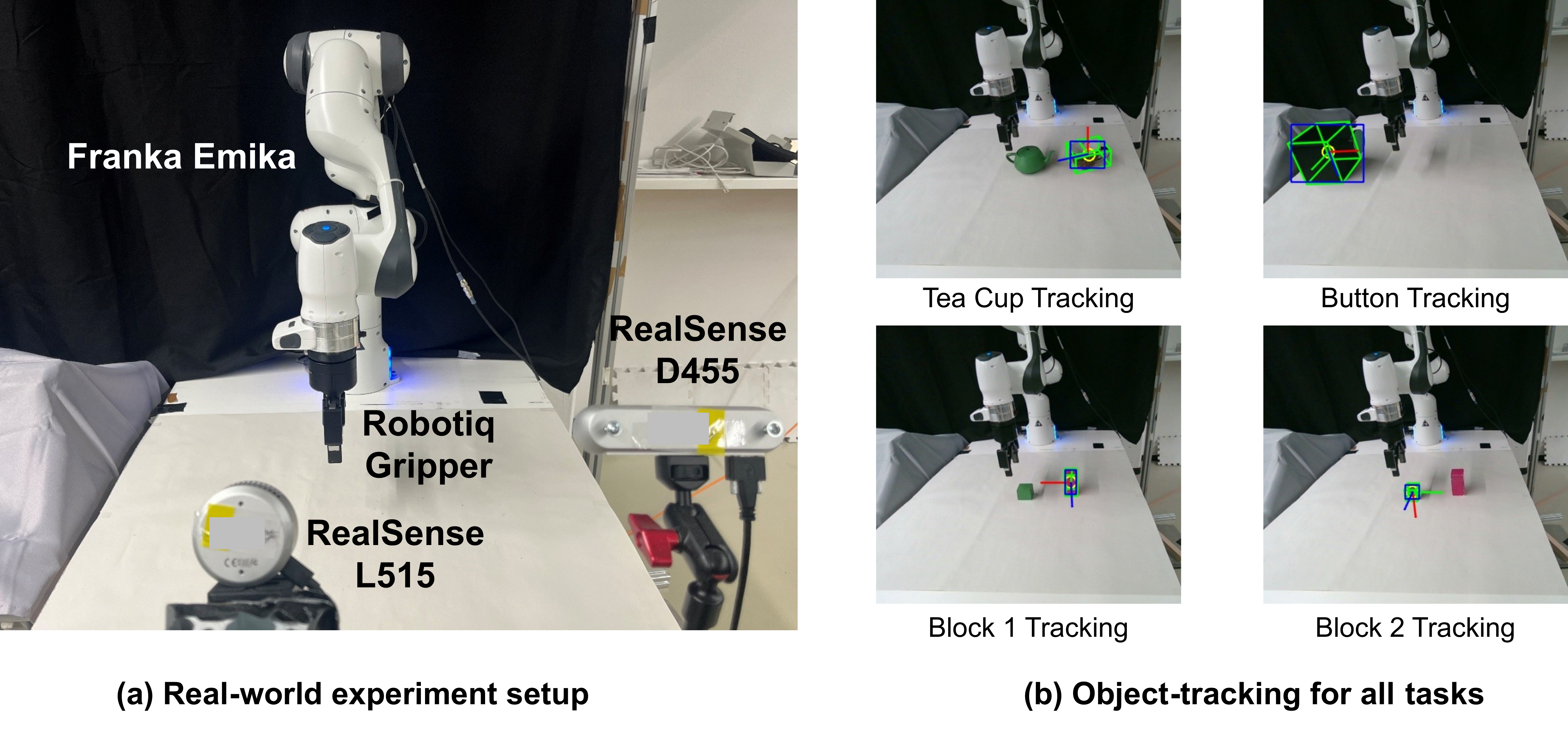}
    \caption{\textbf{Robot Setup and Objects Tracking.} (a) is all the hardware for the real-world experiment. (b) shows the bounding box for object tracking across all tasks.}
  \label{fig:env}
\end{figure}

\noindent \textbf{Hardware and State Setup}
We deploy our framework on a Franka Emika Panda with a Robotiq 2F-85 gripper. An Intel RealSense L515 captures visual observations, while a D455 camera continuously tracks objects. To ensure low latency, raw point clouds are downsampled via Farthest Point Sampling to 2048 points. Hardware setup is shown in \cref{fig:env}. The proprioceptive state consists of the end-effector pose. The action space is formulated as 7-DoF relative commands ($\Delta x, \Delta y, \Delta z$, roll, pitch, yaw) resolved via the \textit{franky} \cite{schneider_franky_2024} inverse kinematics library, plus a binary gripper state (Close or Open).

\noindent \textbf{Data Collection, Network Architecture and Compute}
We collect 25 expert demonstrations per task via a scripted kinematic oracle that injects spatial variance to efficiently mimic human data distributions. For high-speed real-time control, we employ ``Simple DDP'', a lightweight architecture inspired by DP3. It utilizes reduced-capacity 1D U-Nets with down-sampling dimensions of $[128, 256, 384]$ for the Diffusion Policy and $[128, 256, 512]$ for the Diffusion World Model. Finally, we retain all sequence hyperparameters from simulation: $M=2$, $H=16$, $N=8$, and $\lambda=1$.
The models are trained for 2000 epochs on a NVIDIA 4090 GPU. During deployment, the DDP architecture is hosted on an NVIDIA 4070 GPU, with inference times of 40 ms for the diffusion policy and 70 ms for the world model. Concurrently, to guarantee robustness in real-time tracking, we utilize the FoundationPose++ \cite{foundationposeplusplus} module, as shown in \cref{fig:env}(b). For further comprehensive detail, please refer to \cref{sec:supp_real_world} in the Appendix.

\begin{table*}[tb]
  \caption{\textbf{Real Robot Evaluation Results (Success Rate \%).} We evaluate the baseline and our Dream Diffusion Policy (DDP) across three tasks (10 trials each) of varying difficulty under In-Distribution (ID) and Out-of-Distribution (OOD) settings.}
  \label{tab:real_robot_eval}
  \centering
  \resizebox{0.6\textwidth}{!}{
  \begin{tabular}{ l | c | c | c | c }
    \toprule
    \textbf{Algorithm / Condition} & \begin{tabular}{@{}c@{}}\textbf{Press Button}\end{tabular} & \begin{tabular}{@{}c@{}}\textbf{Pour Tea}\end{tabular} & \begin{tabular}{@{}c@{}}\textbf{Stack Blocks}\end{tabular} & \textbf{Average} \\
    \midrule
    \multicolumn{5}{l}{\textbf{Base Policy (ID)}} \\
    \midrule
    DP3 & 100 & 90 & 90 & 93.3 \\
    \textbf{DDP (Ours)} & 100 & 90 & 80 & \textbf{90.0} \\
    \midrule
    \multicolumn{5}{l}{\textbf{Tracking-Augmented (OOD)}} \\
    \midrule
    DP3 (Ablation) & 10 & 0 & 0 & 3.3 \\
    \textbf{DDP (Ours)} & 100 & 90 & 60 & \textbf{83.3} \\
    \midrule
    \multicolumn{5}{l}{\textbf{Imagination Intervention (ID)}} \\
    \midrule
    \textbf{DDP (Open-loop Imagination)} & 100 & 80 & 50 & \textbf{76.7} \\
    \bottomrule
  \end{tabular}
  }
\end{table*}

\begin{figure}[!tb]
  \centering
  \includegraphics[height=2.55cm]{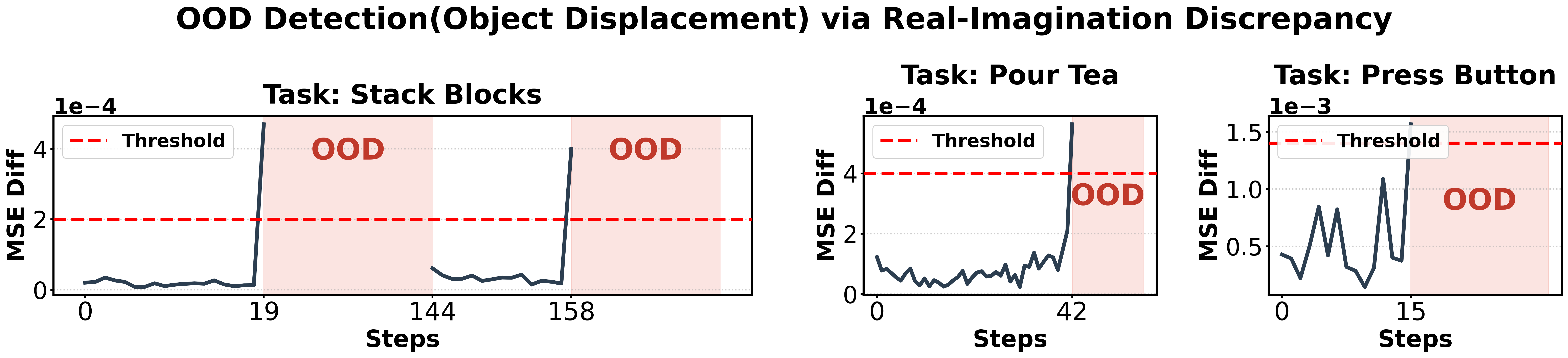}
    \caption{\textbf{OOD Detection via Real-Imagination Discrepancy.} The detector successfully identifies anomalies when objects are accidentally shifted across three different tasks. An OOD state is triggered (shaded regions) when the latent discrepancy ($\mathcal{D}_{R-I}$) exceeds a task-specific threshold $\tau_{diff}$. Specifically, we set $\tau_{diff} = 0.0002$ for Stack Blocks, $\tau_{diff} = 0.0004$ for Pour Tea, and $\tau_{diff} = 0.0012$ for Press Button.}
  \label{fig:loss}

  
  \includegraphics[height=11cm]{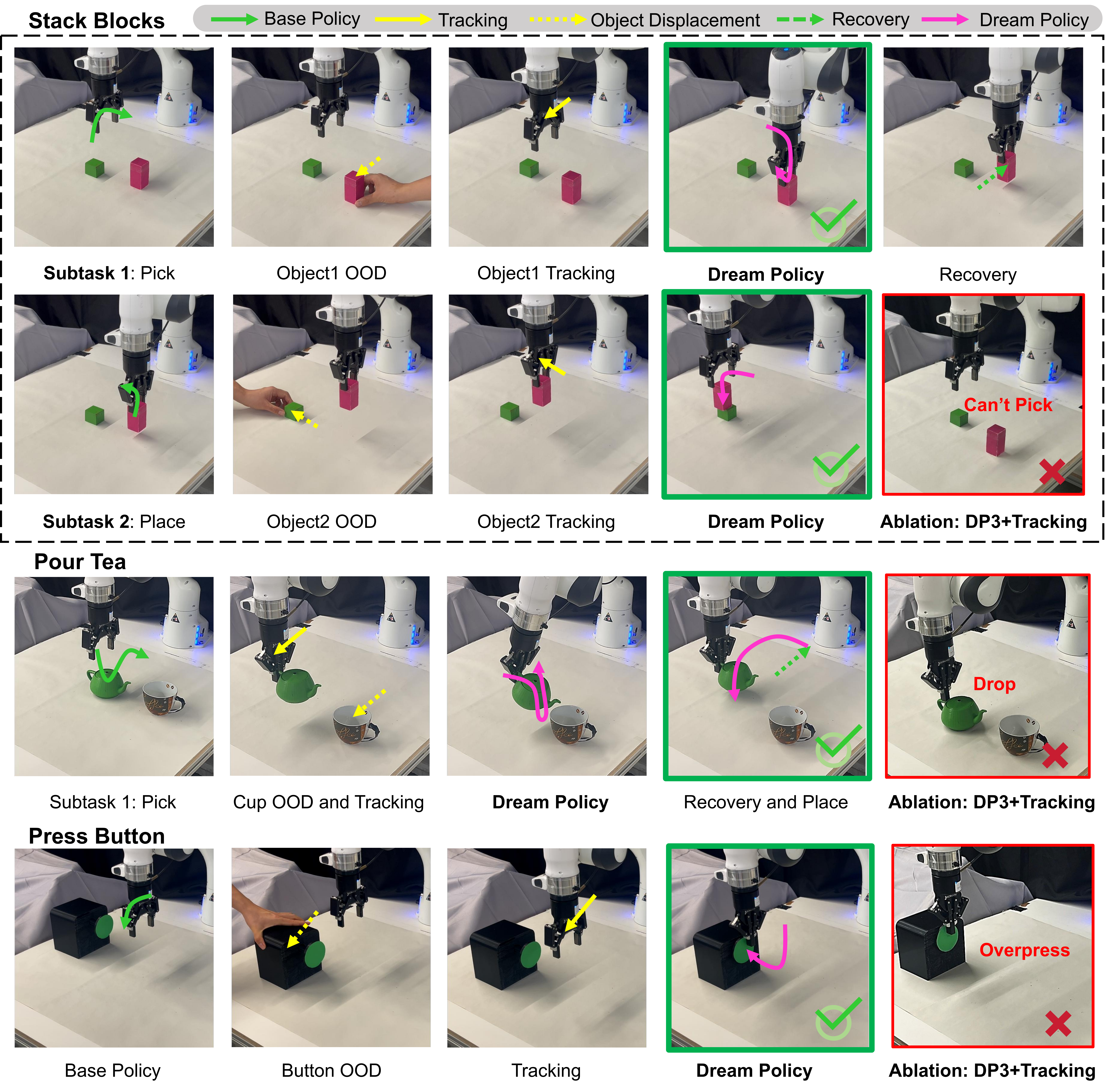}
  \caption{\textbf{Qualitative real-robot execution on Stack Blocks, Pour Tea, and Press Button tasks.} DDP successfully tracks OOD objects, finishes the main actions of subtasks via the imagined Dream Policy, and executes recover to achieve the remaining task success. In contrast, the tracking-augmented DP3 ablation fails across severe OOD scenarios.}
  \label{fig:real_robot_qualitative}
  
\end{figure}

\noindent \textbf{Results} \cref{fig:sumplementary_id} in the Appendix illustrates the successful execution sequences of DDP under In-Distribution (ID) settings for the three real-world tasks: Press Button, Pour Tea, and Stack Blocks. To evaluate OOD robustness, target objects are abruptly moved completely outside the In-Distribution (ID) zone mid-task. As shown in \cref{fig:loss}, our OOD detector identifies these displacements by monitoring \textit{latent discrepancy} in real-time, which quantifies the latent distance between actual visual observations and the policy's imagined projections. Sudden displacements cause the observed visual state to sharply deviate from the imagined trajectory, spiking the discrepancy above a threshold to reliably flag the OOD condition.

\begin{figure}[tb]
  \centering
  \includegraphics[height=5.8cm]{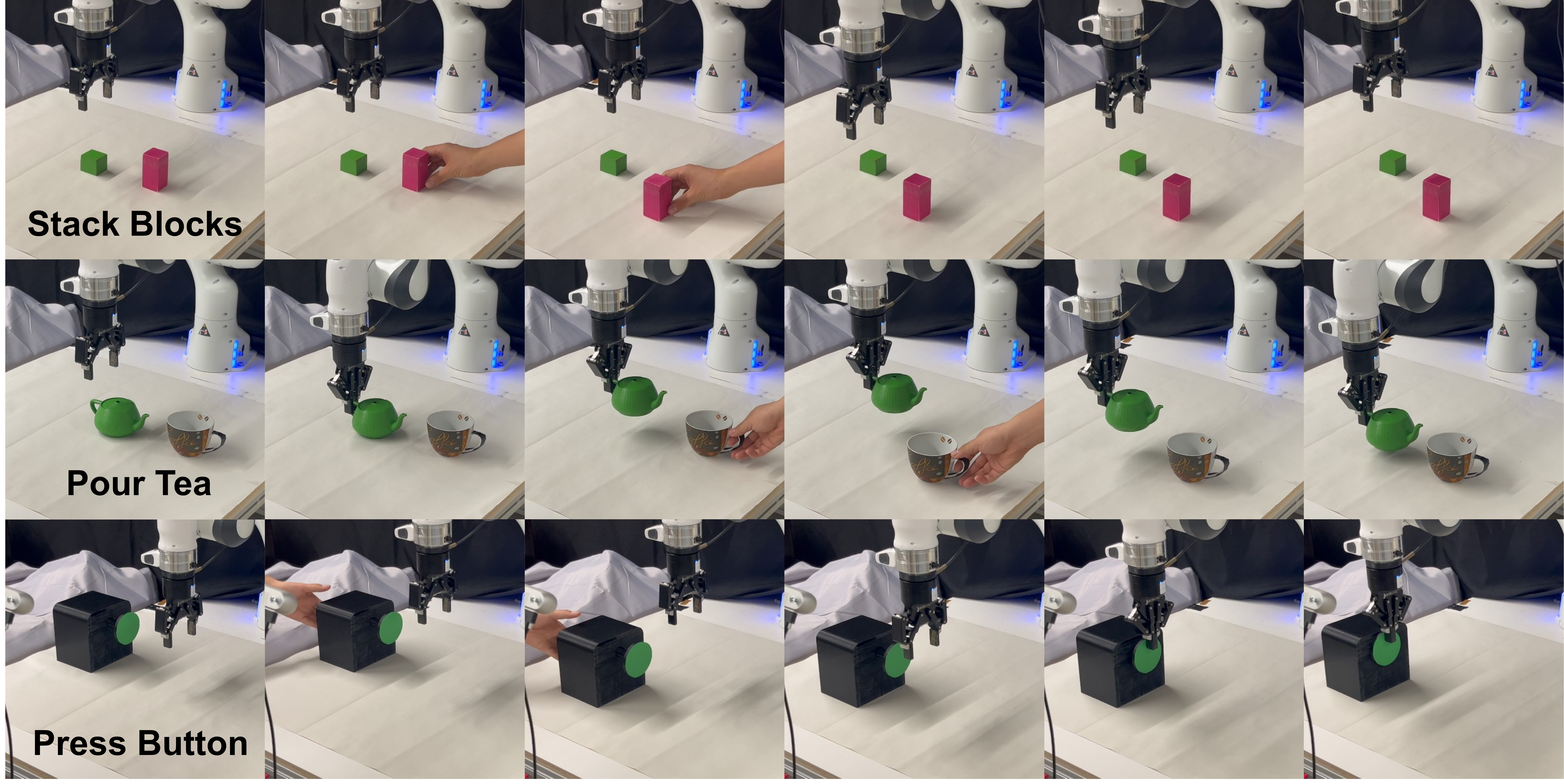}
    \caption{\textbf{Failure modes of the DP3 tracking ablation.} Real-world execution sequences demonstrating how the baseline policy fails under severe Out-of-Distribution (OOD) spatial shifts, resulting in missing grasps, dropping objects, and imprecise pressing (pressing the box after pressing the button).}
  \label{fig:sumplementary_ood_ablation}
\end{figure}

As shown in \cref{tab:real_robot_eval}, our tracking-augmented DDP achieves an 83.3\% average success rate, drastically outperforming the DP3 tracking ablation (3.3\%). \Cref{fig:real_robot_qualitative} illustrates these execution differences. In the two-subtask \textit{Stack Blocks} sequence, DDP tracks OOD shifts for each object, finishes the remaining motions via imagination, and triggers a precise physical recovery between subtasks based on the gripper stage; conversely, the DP3 ablation entirely fails to pick the displaced block. In \textit{Pour Tea}, DDP tracks the shifted cup and seamlessly executes pouring actions from imagination after picking the teapot, whereas DP3 accidentally drops the pot. For \textit{Press Button}, DDP recovers to press the target precisely to its threshold, while the DP3 ablation fails by either under-pressing or over-pressing. Detailed failure mode illustration for each real-world task is provided \cref{fig:sumplementary_ood_ablation}.

Furthermore, we test DDP's reliance on its internal world model by executing tasks using 100\% imagined latents after the first observation chunk (Open-loop Imagination). As detailed in \cref{tab:real_robot_eval}, DDP maintains a remarkable 76.7\% average success rate. This performance confirms that the network's imagined representations are stable and reliable for continuous real-world control.

\begin{figure}[tb]
  \centering
  \includegraphics[height=2.8cm]{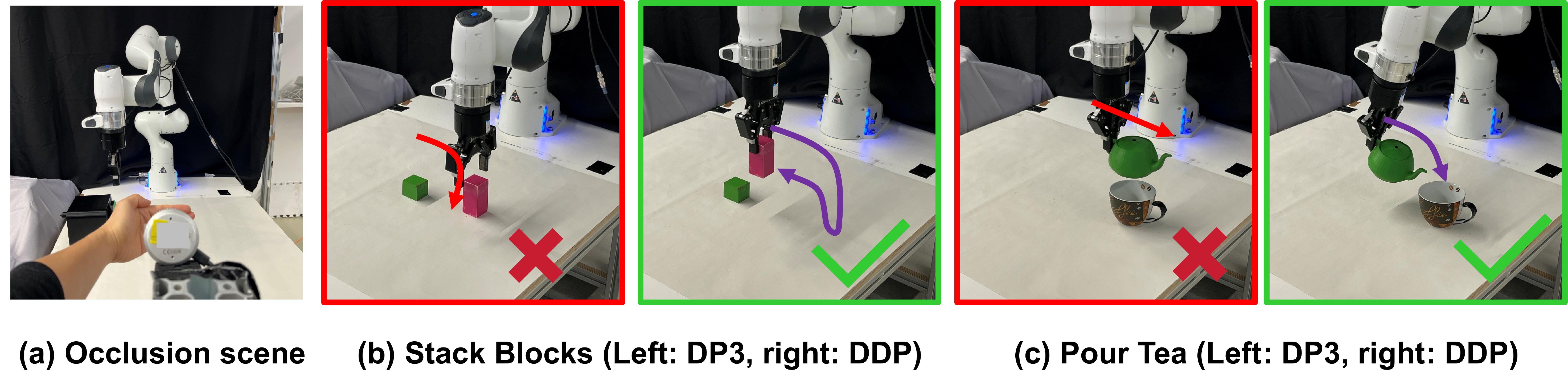}
    \caption{\textbf{Qualitative results under visual occlusion.} (a) A hand obstructs the camera mid-task. In (b) Stack Blocks and (c) Pour Tea, baseline DP3 fails due to visual corruption, whereas DDP seamlessly bypasses the occlusion using internal imagination.}
  \label{fig:occlusion}
\end{figure}

\textbf{Robustness to Visual Occlusion.} To further evaluate DDP against severe visual occlusion, we block the camera mid-task. This creates a massive real-imagination discrepancy ($\mathcal{D}_{R-I}$), instantly triggering OOD mode. While the DP3 baseline fails when blinded, DDP safely bypasses the corruption to complete both \textit{Stack Blocks} and \textit{Pour Tea} purely via autoregressive imagination (\cref{fig:occlusion}).

\section{Conclusion}
\label{sec:conclusion}
We introduced the Dream Diffusion Policy (DDP) to enhance visuomotor robustness against severe out-of-distribution (OOD) disturbances. Co-optimizing a diffusion policy and world model via a shared 3D encoder embeds predictive dynamics directly into DDP's representation. During inference, DDP detects OOD anomalies via a real-imagination discrepancy ($\mathcal{D}_{R-I}$). Rather than collapsing under corrupted vision, it safely bypasses these shifts using an autoregressive imagination loop—executing actions from latent forecasts before triggering a localized physical recovery. Evaluations confirm DDP significantly outperforms baselines \textbf{specifically in OOD scenarios}, maintaining a high success rate even under almost complete visual blindness where standard methods fail.

\textbf{Limitations and Future Work.} DDP has three main limitations: (1) It highly relies on external tracking modules, allowing drift to accumulate across multi-stage subtasks, (2) tasks must begin strictly In-Distribution (ID) to properly anchor the initial observation, and (3) the inference mechanism struggles with compounding OOD events or low-level action disruptions (e.g., gripper slippage). Addressing these dependencies remains a key direction for future research. Potential solutions, such as employing meta-reinforcement learning~\cite{bing2023meta,wang2023metareinforcement,yao2023learning} to adaptively fine-tune the world model online, integrating the lifelong learning mechanism~\cite{meng2025preserving} to preserve and combine prior knowledge, or integrating low-level tactile feedback~\cite{wu2025tacdiffusion}, could further enhance DDP's resilience in real-world applications. We also plan to extend DDP to more complex multi-object manipulation tasks and explore its integration with language-conditioned methods~\cite{zhou2023languageconditioned,10934975} or large language models for higher-level reasoning and planning~\cite{yao2023bridging}.


%
%
\bibliographystyle{splncs04}
\bibliography{main}

@String(ICLR  = {Int. Conf. Learn. Represent.})

@String(AAAI  = {AAAI})

@String(ICLR  = {ICLR})

@article{ho2020denoising,
  title={Denoising diffusion probabilistic models},
  author={Ho, Jonathan and Jain, Ajay and Abbeel, Pieter},
  journal={Advances in neural information processing systems},
  volume={33},
  pages={6840--6851},
  year={2020}
}

@article{chi2025diffusion,
  title={Diffusion policy: Visuomotor policy learning via action diffusion},
  author={Chi, Cheng and Xu, Zhenjia and Feng, Siyuan and Cousineau, Eric and Du, Yilun and Burchfiel, Benjamin and Tedrake, Russ and Song, Shuran},
  journal={The International Journal of Robotics Research},
  volume={44},
  number={10-11},
  pages={1684--1704},
  year={2025},
  publisher={Sage Publications Sage UK: London, England}
}

@ARTICLE{10855557,
  author={Seo, Mingyo and Park, H. Andy and Yuan, Shenli and Zhu, Yuke and Sentis, Luis},
  journal={IEEE Robotics and Automation Letters}, 
  title={LEGATO: Cross-Embodiment Imitation Using a Grasping Tool}, 
  year={2025},
  volume={10},
  number={3},
  pages={2854-2861},
  doi={10.1109/LRA.2025.3535182}}

@inproceedings{zhang2025flowpolicy,
  title={Flowpolicy: Enabling fast and robust 3d flow-based policy via consistency flow matching for robot manipulation},
  author={Zhang, Qinglun and Liu, Zhen and Fan, Haoqiang and Liu, Guanghui and Zeng, Bing and Liu, Shuaicheng},
  booktitle={Proceedings of the AAAI Conference on Artificial Intelligence},
  volume={39},
  number={14},
  pages={14754--14762},
  year={2025}
}

@inproceedings{perez2018film,
  title={Film: Visual reasoning with a general conditioning layer},
  author={Perez, Ethan and Strub, Florian and De Vries, Harm and Dumoulin, Vincent and Courville, Aaron},
  booktitle={Proceedings of the AAAI conference on artificial intelligence},
  volume={32},
  number={1},
  year={2018}
}

@article{wang2025hierarchical,
  title={Hierarchical diffusion policy: manipulation trajectory generation via contact guidance},
  author={Wang, Dexin and Liu, Chunsheng and Chang, Faliang and Xu, Yichen},
  journal={IEEE Transactions on Robotics},
  year={2025},
  publisher={IEEE}
}

@article{ding2024diffusion,
  title={Diffusion world model: Future modeling beyond step-by-step rollout for offline reinforcement learning},
  author={Ding, Zihan and Zhang, Amy and Tian, Yuandong and Zheng, Qinqing},
  journal={arXiv preprint arXiv:2402.03570},
  year={2024}
}

@inproceedings{saharia2022palette,
  title={Palette: Image-to-image diffusion models},
  author={Saharia, Chitwan and Chan, William and Chang, Huiwen and Lee, Chris and Ho, Jonathan and Salimans, Tim and Fleet, David and Norouzi, Mohammad},
  booktitle={ACM SIGGRAPH 2022 conference proceedings},
  pages={1--10},
  year={2022}
}

@inproceedings{ma2024hierarchical,
  title={Hierarchical diffusion policy for kinematics-aware multi-task robotic manipulation},
  author={Ma, Xiao and Patidar, Sumit and Haughton, Iain and James, Stephen},
  booktitle={Proceedings of the IEEE/CVF Conference on Computer Vision and Pattern Recognition},
  pages={18081--18090},
  year={2024}
}

@article{geirhos2020shortcut,
  title={Shortcut learning in deep neural networks},
  author={Geirhos, Robert and Jacobsen, J{\"o}rn-Henrik and Michaelis, Claudio and Zemel, Richard and Brendel, Wieland and Bethge, Matthias and Wichmann, Felix A},
  journal={Nature Machine Intelligence},
  volume={2},
  number={11},
  pages={665--673},
  year={2020},
  publisher={Nature Publishing Group UK London}
}

@inproceedings{xue2025reactive,
  title     = {Reactive Diffusion Policy: Slow-Fast Visual-Tactile Policy Learning for Contact-Rich Manipulation},
  author    = {Xue, Han and Ren, Jieji and Chen, Wendi and Zhang, Gu and Fang, Yuan and Gu, Guoying and Xu, Huazhe and Lu, Cewu},
  booktitle = {Proceedings of Robotics: Science and Systems (RSS)},
  year      = {2025}
}

@ARTICLE{10912754,
  author={Wang, Dexin and Liu, Chunsheng and Chang, Faliang and Xu, Yichen},
  journal={IEEE Transactions on Robotics}, 
  title={Hierarchical Diffusion Policy: Manipulation Trajectory Generation via Contact Guidance}, 
  year={2025},
  volume={41},
  number={},
  pages={2086-2104},
  doi={10.1109/TRO.2025.3547272}}

@inproceedings{tobin2017domain,
  title={Domain randomization for transferring deep neural networks from simulation to the real world},
  author={Tobin, Joshua and Fong, Rachel and Ray, Alex and Schneider, Jonas and Zaremba, Wojciech and Abbeel, Pieter},
  booktitle={2017 IEEE/RSJ International Conference on Intelligent Robots and Systems (IROS)},
  pages={23--30},
  year={2017},
  organization={IEEE}
}

@article{yao2025inference,
  title={Inference-stage Adaptation-projection Strategy Adapts Diffusion Policy to Cross-manipulators Scenarios},
  author={Yao, Xiangtong and Zhou, Yirui and Meng, Yuan and Liu, Yanwen and Dong, Liangyu and Zhang, Zitao and Bing, Zhenshan and Huang, Kai and Sun, Fuchun and Knoll, Alois},
  journal={arXiv preprint arXiv:2509.11621},
  year={2025}
}

@inproceedings{chandra2025diwa,
  title={DiWA: Diffusion Policy Adaptation with World Models},
  author={Chandra, Akshay L and Nematollahi, Iman and Huang, Chenguang and Welschehold, Tim and Burgard, Wolfram and Valada, Abhinav},
  booktitle={Conference on Robot Learning},
  pages={3378--3400},
  year={2025},
  organization={PMLR}
}

@inproceedings{wang2020tent,
  title={{Tent}: Fully Test-Time Adaptation by Entropy Minimization},
  author={Wang, Dequan and Shelhamer, Evan and Liu, Shaoteng and Olshausen, Bruno and Darrell, Trevor},
  booktitle={International Conference on Learning Representations (ICLR)},
  year={2021}
}

@article{ha2018world,
  title={World models},
  author={Ha, David and Schmidhuber, J{\"u}rgen},
  journal={arXiv preprint arXiv:1803.10122},
  year={2018}
}

@inproceedings{Ze2024DP3,
	title={3D Diffusion Policy: Generalizable Visuomotor Policy Learning via Simple 3D Representations},
	author={Yanjie Ze and Gu Zhang and Kangning Zhang and Chenyuan Hu and Muhan Wang and Huazhe Xu},
	booktitle={Proceedings of Robotics: Science and Systems (RSS)},
	year={2024}
}

@article{ada2024diffusion,
  title={Diffusion policies for out-of-distribution generalization in offline reinforcement learning},
  author={Ada, Suzan Ece and Oztop, Erhan and Ugur, Emre},
  journal={IEEE Robotics and Automation Letters},
  volume={9},
  number={4},
  pages={3116--3123},
  year={2024},
  publisher={IEEE}
}

@inproceedings{
du2025dynaguide,
title={DynaGuide: Steering Diffusion Polices with Active Dynamic Guidance},
author={Max Du and Shuran Song},
booktitle={The Thirty-ninth Annual Conference on Neural Information Processing Systems},
year={2025},
url={https://openreview.net/forum?id=XOw7Yf8qN3}
}

@inproceedings{
curtis2025flowbased,
title={Flow-based Domain Randomization for Learning and Sequencing Robotic Skills},
author={Aidan Curtis and Eric Li and Michael Noseworthy and Nishad Gothoskar and Sachin Chitta and Hui Li and Leslie Pack Kaelbling and Nicole E Carey},
booktitle={Forty-second International Conference on Machine Learning},
year={2025},
url={https://openreview.net/forum?id=9JQXuyzdGL}
}

@article{huang2025improving,
  title={Improving Robustness to Out-of-Distribution States in Imitation Learning via Deep Koopman-Boosted Diffusion Policy},
  author={Huang, Dianye and Navab, Nassir and Jiang, Zhongliang},
  journal={IEEE Transactions on Robotics},
  volume={41},
  pages={6680--6692},
  year={2025},
  publisher={IEEE}
}

@inproceedings{tian2025pdfactor,
  title={Pdfactor: Learning tri-perspective view policy diffusion field for multi-task robotic manipulation},
  author={Tian, Jingyi and Wang, Le and Zhou, Sanping and Wang, Sen and Li, Jiayi and Sun, Haowen and Tang, Wei},
  booktitle={Proceedings of the Computer Vision and Pattern Recognition Conference},
  pages={15757--15767},
  year={2025}
}

@inproceedings{huang2024diffusion,
  title={Diffusion reward: Learning rewards via conditional video diffusion},
  author={Huang, Tao and Jiang, Guangqi and Ze, Yanjie and Xu, Huazhe},
  booktitle={European Conference on Computer Vision},
  pages={478--495},
  year={2024},
  organization={Springer}
}

@inproceedings{wu2025device,
  title={On-device diffusion transformer policy for efficient robot manipulation},
  author={Wu, Yiming and Wang, Huan and Chen, Zhenghao and Pang, Jianxin and Xu, Dong},
  booktitle={Proceedings of the IEEE/CVF International Conference on Computer Vision},
  pages={14073--14083},
  year={2025}
}

@inproceedings{wu2025afforddp,
  title={Afforddp: Generalizable diffusion policy with transferable affordance},
  author={Wu, Shijie and Zhu, Yihang and Huang, Yunao and Zhu, Kaizhen and Gu, Jiayuan and Yu, Jingyi and Shi, Ye and Wang, Jingya},
  booktitle={Proceedings of the Computer Vision and Pattern Recognition Conference},
  pages={6971--6980},
  year={2025}
}

@inproceedings{gao2025out,
  title={Out-of-distribution recovery with object-centric keypoint inverse policy for visuomotor imitation learning},
  author={Gao, George Jiayuan and Li, Tianyu and Figueroa, Nadia},
  booktitle={2025 IEEE/RSJ International Conference on Intelligent Robots and Systems (IROS)},
  pages={9816--9828},
  year={2025},
  organization={IEEE}
}

@article{he2025demystifying,
  title={Demystifying diffusion policies: Action memorization and simple lookup table alternatives},
  author={He, Chengyang and Liu, Xu and Camps, Gadiel Sznaier and Sartoretti, Guillaume and Schwager, Mac},
  journal={arXiv preprint arXiv:2505.05787},
  year={2025}
}

@inproceedings{yu2020meta,
  title={Meta-world: A benchmark and evaluation for multi-task and meta reinforcement learning},
  author={Yu, Tianhe and Quillen, Deirdre and He, Zhanpeng and Julian, Ryan and Hausman, Karol and Finn, Chelsea and Levine, Sergey},
  booktitle={Conference on robot learning},
  pages={1094--1100},
  year={2020},
  organization={PMLR}
}

@inproceedings{wu2025tacdiffusion,
  title={Tacdiffusion: Force-domain diffusion policy for precise tactile manipulation},
  author={Wu, Yansong and Chen, Zongxie and Wu, Fan and Chen, Lingyun and Zhang, Liding and Bing, Zhenshan and Swikir, Abdalla and Haddadin, Sami and Knoll, Alois},
  booktitle={2025 IEEE International Conference on Robotics and Automation (ICRA)},
  pages={11831--11837},
  year={2025},
  organization={IEEE}
}

@article{meng2025preserving,
  title={Preserving and combining knowledge in robotic lifelong reinforcement learning},
  author={Meng, Yuan and Bing, Zhenshan and Yao, Xiangtong and Chen, Kejia and Huang, Kai and Gao, Yang and Sun, Fuchun and Knoll, Alois},
  journal={Nature Machine Intelligence},
  volume={7},
  number={2},
  pages={256--269},
  year={2025},
  publisher={Nature Publishing Group UK London}
}

@article{yao2023bridging,
  title={Bridging language and action: A survey of language-conditioned robot manipulation},
  author={Yao, Xiangtong and Zhou, Hongkuan and Mees, Oier and Meng, Yuan and Xiao, Ted and Bisk, Yonatan and Oh, Jean and Johns, Edward and Shridhar, Mohit and Shah, Dhruv and others},
  journal={arXiv preprint arXiv:2312.10807},
  year={2023}
}

@ARTICLE{10934975,
  author={Yao, Xiangtong and Blei, Tobias and Meng, Yuan and Zhang, Yu and Zhou, Hongkuan and Bing, Zhenshan and Huang, Kai and Sun, Fuchun and Knoll, Alois},
  journal={IEEE/ASME Transactions on Mechatronics}, 
  title={Long-Horizon Language-Conditioned Imitation Learning for Robotic Manipulation}, 
  year={2025},
  volume={},
  number={},
  pages={1-12},}

@article{zhou2023languageconditioned,
  author={Hongkuan Zhou and Zhenshan Bing and Xiangtong Yao and Xiaojie Su and Chenguang Yang and Kai Huang and Alois Knoll},
  journal={IEEE Robotics and Automation Letters}, 
  title={Language-Conditioned Imitation Learning With Base Skill Priors Under Unstructured Data}, 
  year={2024},
  volume={9},
  number={11},
  pages={9805-9812},
}

@inproceedings{bing2023meta,
  title={Meta-reinforcement learning via language instructions},
  author={Bing, Zhenshan and Koch, Alexander and Yao, Xiangtong and Huang, Kai and Knoll, Alois},
  booktitle={2023 IEEE International Conference on Robotics and Automation (ICRA)},
  pages={5985--5991},
  year={2023},
  organization={IEEE}
}

@inproceedings{yao2023learning,
  title={Learning from symmetry: Meta-reinforcement learning with symmetrical behaviors and language instructions},
  author={Yao, Xiangtong and Bing, Zhenshan and Zhuang, Genghang and Chen, Kejia and Zhou, Hongkuan and Huang, Kai and Knoll, Alois},
  booktitle={IEEE/RSJ International Conference on Intelligent Robots and Systems (IROS)},
  pages={5574--5581},
  year={2023},
  organization={IEEE}
}

@inproceedings{wang2023metareinforcement,
  title={Meta-reinforcement learning based on self-supervised task representation learning},
  author={Wang, Mingyang and Bing, Zhenshan and Yao, Xiangtong and Wang, Shuai and Kai, Huang and Su, Hang and Yang, Chenguang and Knoll, Alois},
  booktitle={Proceedings of the AAAI Conference on Artificial Intelligence},
  volume={37},
  number={8},
  pages={10157--10165},
  year={2023}
}

@article{yao2025pick,
  title={Pick-and-place manipulation across grippers without retraining: A learning-optimization diffusion policy approach},
  author={Yao, Xiangtong and Zhou, Yirui and Meng, Yuan and Dong, Liangyu and Hong, Lin and Zhang, Zitao and Bing, Zhenshan and Huang, Kai and Sun, Fuchun and Knoll, Alois},
  journal={arXiv preprint arXiv:2502.15613},
  year={2025}
}

@article{rajeswaran2017learning,
  title={Learning complex dexterous manipulation with deep reinforcement learning and demonstrations},
  author={Rajeswaran, Aravind and Kumar, Vikash and Gupta, Abhishek and Vezzani, Giulia and Schulman, John and Todorov, Emanuel and Levine, Sergey},
  journal={arXiv preprint arXiv:1709.10087},
  year={2017}
}

@inproceedings{wen2024foundationpose,
  title={Foundationpose: Unified 6d pose estimation and tracking of novel objects},
  author={Wen, Bowen and Yang, Wei and Kautz, Jan and Birchfield, Stan},
  booktitle={Proceedings of the IEEE/CVF conference on computer vision and pattern recognition},
  pages={17868--17879},
  year={2024}
}

@misc{foundationposeplusplus,
  author       = {Wenhao Yan and Jie Chu},
  title        = {FoundationPose++: Simple Tricks Boost FoundationPose Performance in High-Dynamic Scenes},
  howpublished = {\url{https://github.com/teal024/FoundationPose-plus-plus}},
  year         = {2025},
  note         = {GitHub Repository}
}

@article{lai2022action,
  title={Action chunking as policy compression},
  author={Lai, Lucy and Huang, Ann Zixiang and Gershman, Samuel J},
  journal={PsyArXiv},
  year={2022}
}

@article{liu2024bidirectional,
  title={Bidirectional decoding: Improving action chunking via guided test-time sampling},
  author={Liu, Yuejiang and Hamid, Jubayer Ibn and Xie, Annie and Lee, Yoonho and Du, Maximilian and Finn, Chelsea},
  journal={arXiv preprint arXiv:2408.17355},
  year={2024}
}

@article{cen2025worldvla,
  title={Worldvla: Towards autoregressive action world model},
  author={Cen, Jun and Yu, Chaohui and Yuan, Hangjie and Jiang, Yuming and Huang, Siteng and Guo, Jiayan and Li, Xin and Song, Yibing and Luo, Hao and Wang, Fan and others},
  journal={arXiv preprint arXiv:2506.21539},
  year={2025}
}

@article{morales2024exponential,
  title={Exponential moving average of weights in deep learning: Dynamics and benefits},
  author={Morales-Brotons, Daniel and Vogels, Thijs and Hendrikx, Hadrien},
  journal={arXiv preprint arXiv:2411.18704},
  year={2024}
}

@article{wang2026palm,
  title={PALM: Enhanced Generalizability for Local Visuomotor Policies via Perception Alignment},
  author={Wang, Ruiyu and Zhuang, Zheyu and Kragic, Danica and Pokorny, Florian T},
  journal={arXiv preprint arXiv:2601.19514},
  year={2026}
}

@article{sun2025latent,
  title={Latent policy barrier: Learning robust visuomotor policies by staying in-distribution},
  author={Sun, Zhanyi and Song, Shuran},
  journal={arXiv preprint arXiv:2508.05941},
  year={2025}
}

@article{zhang2024diffusion,
  title={Diffusion meets dagger: Supercharging eye-in-hand imitation learning},
  author={Zhang, Xiaoyu and Chang, Matthew and Kumar, Pranav and Gupta, Saurabh},
  journal={arXiv preprint arXiv:2402.17768},
  year={2024}
}

@inproceedings{zhu2023viola,
  title={Viola: Imitation learning for vision-based manipulation with object proposal priors},
  author={Zhu, Yifeng and Joshi, Abhishek and Stone, Peter and Zhu, Yuke},
  booktitle={Conference on Robot Learning},
  pages={1199--1210},
  year={2023},
  organization={PMLR}
}

@article{isaku2025out,
  title={Out of Distribution Detection in Self-adaptive Robots with AI-powered Digital Twins},
  author={Isaku, Erblin and Sartaj, Hassan and Ali, Shaukat and Sanguino, Beatriz and Wang, Tongtong and Li, Guoyuan and Zhang, Houxiang and Peyrucain, Thomas},
  journal={arXiv preprint arXiv:2509.12982},
  year={2025}
}

@inproceedings{graham2023denoising,
  title={Denoising diffusion models for out-of-distribution detection},
  author={Graham, Mark S and Pinaya, Walter HL and Tudosiu, Petru-Daniel and Nachev, Parashkev and Ourselin, Sebastien and Cardoso, Jorge},
  booktitle={Proceedings of the IEEE/CVF conference on computer vision and pattern recognition},
  pages={2948--2957},
  year={2023}
}

@inproceedings{lee2025diff,
  title={Diff-dagger: Uncertainty estimation with diffusion policy for robotic manipulation},
  author={Lee, Sung-Wook and Kang, Xuhui and Kuo, Yen-Ling},
  booktitle={2025 IEEE International Conference on Robotics and Automation (ICRA)},
  pages={4845--4852},
  year={2025},
  organization={IEEE}
}

@article{alonso2024diffusion,
  title={Diffusion for world modeling: Visual details matter in atari},
  author={Alonso, Eloi and Jelley, Adam and Micheli, Vincent and Kanervisto, Anssi and Storkey, Amos J and Pearce, Tim and Fleuret, Fran{\c{c}}ois},
  journal={Advances in Neural Information Processing Systems},
  volume={37},
  pages={58757--58791},
  year={2024}
}

@inproceedings{seo2023masked,
  title={Masked world models for visual control},
  author={Seo, Younggyo and Hafner, Danijar and Liu, Hao and Liu, Fangchen and James, Stephen and Lee, Kimin and Abbeel, Pieter},
  booktitle={Conference on Robot Learning},
  pages={1332--1344},
  year={2023},
  organization={PMLR}
}

@article{ding2025understanding,
  title={Understanding world or predicting future? a comprehensive survey of world models},
  author={Ding, Jingtao and Zhang, Yunke and Shang, Yu and Zhang, Yuheng and Zong, Zefang and Feng, Jie and Yuan, Yuan and Su, Hongyuan and Li, Nian and Sukiennik, Nicholas and others},
  journal={ACM Computing Surveys},
  volume={58},
  number={3},
  pages={1--38},
  year={2025},
  publisher={ACM New York, NY}
}

@article{agarwal2025cosmos,
  title={Cosmos world foundation model platform for physical ai},
  author={Agarwal, Niket and Ali, Arslan and Bala, Maciej and Balaji, Yogesh and Barker, Erik and Cai, Tiffany and Chattopadhyay, Prithvijit and Chen, Yongxin and Cui, Yin and Ding, Yifan and others},
  journal={arXiv preprint arXiv:2501.03575},
  year={2025}
}

@article{hafner2019dream,
  title={Dream to control: Learning behaviors by latent imagination},
  author={Hafner, Danijar and Lillicrap, Timothy and Ba, Jimmy and Norouzi, Mohammad},
  journal={arXiv preprint arXiv:1912.01603},
  year={2019}
}

@article{hafner2020mastering,
  title={Mastering atari with discrete world models},
  author={Hafner, Danijar and Lillicrap, Timothy and Norouzi, Mohammad and Ba, Jimmy},
  journal={arXiv preprint arXiv:2010.02193},
  year={2020}
}

@article{hafner2023mastering,
  title={Mastering diverse domains through world models},
  author={Hafner, Danijar and Pasukonis, Jurgis and Ba, Jimmy and Lillicrap, Timothy},
  journal={arXiv preprint arXiv:2301.04104},
  year={2023}
}

@article{ye2026world,
  title={World Action Models are Zero-shot Policies},
  author={Ye, Seonghyeon and Ge, Yunhao and Zheng, Kaiyuan and Gao, Shenyuan and Yu, Sihyun and Kurian, George and Indupuru, Suneel and Tan, You Liang and Zhu, Chuning and Xiang, Jiannan and others},
  journal={arXiv preprint arXiv:2602.15922},
  year={2026}
}

@software{schneider_franky_2024,
  author       = {Schneider, Tim},
  title        = {franky: High-Level Control Library for Franka Robots},
  year         = {2024},
  publisher    = {GitHub},
  journal      = {GitHub repository},
  howpublished = {\url{https://github.com/TimSchneider42/franky}},
  note         = {Version 1.1.3} 
}

\section*{Appendix}
\setcounter{subsection}{0}
\renewcommand{\thesubsection}{\Alph{subsection}}
\renewcommand{\thesubsubsection}{\thesubsection\arabic{subsubsection}}

\subsection{Pseudocode for Closed-Loop Inference}
\label{sec:supp_inference_algo}

To further clarify the transition between physical reality (In-Distribution) and internal imagination (Out-of-Distribution), we provide the detailed pseudocode for the DDP inference stage in \cref{alg:inference_stage}. The algorithm highlights the real-time detection via latent discrepancy $\mathcal{D}_{R-I}$, the recursive action generation during the OOD state, and the target-specific recovery mechanism triggered at the end of each subtask.

\begin{algorithm}[h]
\caption{DDP Inference with Target-Specific Recovery}
\label{alg:inference_stage}
\KwIn{Visual Encoder $E_\psi$, Policy $\pi_\theta$, World Model $\mathcal{W}_\phi$, Pose Estimator}
\KwIn{Subtasks $\mathcal{S}=\{S_1, \dots, S_J\}$, Expected ID poses $\mathbf{p}^{exp}$, Threshold $\tau_{diff}$}

$\text{OOD\_Mode} \leftarrow \text{False}, \ \mathcal{C}_{disp} \leftarrow \emptyset, \ \mathbf{O}_{curr} \leftarrow E_\psi(X_{real}^0)$\;

\For{subtask $S_j \in \mathcal{S}$ targeting object $C_j$}{
    \While{completion trigger $\mathcal{T}_{end}(S_j)$ is \textbf{False}}{
        $\mathbf{O}_{real}^t \leftarrow E_\psi(X_{real}^t)$\;
        
        \tcp{1. OOD Detection \& Tracking}
        \If{\textbf{not} $\text{OOD\_Mode}$}{
            $\mathbf{O}_{pred}^t \leftarrow \mathcal{W}_\phi(\mathbf{O}_{real}^{t-1}, \mathbf{a}_{t-1})$\;
            $\mathcal{D}_{R-I}(t) \leftarrow \| \mathbf{O}_{real}^t - \mathbf{O}_{pred}^t \|_2^2$\;
            
            \If{$\mathcal{D}_{R-I}(t) > \tau_{diff}$}{
                $\text{OOD\_Mode} \leftarrow \text{True}, \ \mathcal{C}_{disp} \leftarrow \text{Identify displaced objects}$\;
                \If{$C_j \in \mathcal{C}_{disp}$}{
                    Execute tracking action $\mathbf{a}^{track}_j = \mathbf{p}_{C_j} - \mathbf{p}_{C_j}^{exp}$\;
                }
                $\mathbf{O}_{curr} \leftarrow \mathbf{O}_{pred}^t$ \tcp*{Enter Imagination}
            }
            \Else{
                $\mathbf{O}_{curr} \leftarrow \mathbf{O}_{real}^t$ \tcp*{Nominal ID Reality}
            }
        }
        
        \tcp{2. Action Execution \& Recursive Imagination}
        $\mathbf{a}_t \sim \pi_\theta(\cdot \mid \mathbf{O}_{curr}), \quad \text{Execute action chunk } \mathbf{a}_t$\;
        \If{$\text{OOD\_Mode}$}{
            $\mathbf{O}_{curr} \leftarrow \mathcal{W}_\phi(\mathbf{O}_{curr}, \mathbf{a}_t)$ \tcp*{Autoregressive Forecast}
        }
        $t \leftarrow t + 1$\;
    }
    
    \tcp{3. Target-Specific Recovery}
    \If{$\text{OOD\_Mode}$ \textbf{and} $C_j \in \mathcal{C}_{disp}$}{
        Execute recovery action $\mathbf{a}^{rec}_j = \mathbf{p}_{C_j} - \mathbf{p}_{C_j}^{exp}$\;
        $\mathcal{C}_{disp} \leftarrow \mathcal{C}_{disp} \setminus \{C_j\}$\;
        
        \tcp{Re-evaluate ID State}
        $\mathcal{D}_{R-I}(t) \leftarrow \| E_\psi(X_{real}^t) - \mathcal{W}_\phi(\mathbf{O}_{curr}, \mathbf{a}_{t-1}) \|_2^2$\;
        
        \If{$\mathcal{D}_{R-I}(t) \le \tau_{diff}$}{
            $\text{OOD\_Mode} \leftarrow \text{False}$ \tcp*{Return to Reality}
        }
    }
}
\end{algorithm}

\begin{table}[ht]
\centering
\caption{\textbf{Detailed OOD Disturbance Parameters for MetaWorld Tasks.} The displacement vector $\Delta \mathbf{P}$ is represented as $[dx, dy, dz]$ in meters.}
\label{tab:sim_ood_details}
\resizebox{\textwidth}{!}{
\begin{tabular}{l l c l c}
\toprule
\textbf{Task Name} & \textbf{Displaced Object(s)} & \textbf{Intervention Timing} & \textbf{Displacement Vector ($\Delta \mathbf{P}$)} & \textbf{Tracking Logic} \\
\midrule
Assembly & Peg (\texttt{peg}) & Chunk 6 & $[-0.25, 0.0, 0.0]$ & Single-Stage \\
Button Press & Box \& Button (\texttt{box}) & Chunk 3 & $[-0.25, 0.0, 0.0]$ & Single-Stage \\
Drawer Open & Drawer (\texttt{drawer}) & Chunk 2 & $[0.0, 0.0, +0.10]$ & Single-Stage \\
Hammer & Box \& Nail (\texttt{box}, \texttt{nailHead}) & Chunk 5 & $[0.0, +0.10, 0.0]$ & Single-Stage \\
Door Unlock & Door (\texttt{door}) & Chunk 2 & $[0.0, +0.10, 0.0]$ & Single-Stage \\
Soccer & Goal Net (\texttt{goal\_whole}) & Chunk 2 & $[-0.20, +0.10, 0.0]$ & Single-Stage \\
\midrule
Pick Place & Puck (\texttt{obj}) & Chunk 2 & $[-0.20, -0.10, 0.0]$ & Two-Stage (+ Safe Lift) \\
Push & Puck (\texttt{obj}) & Chunk 2 & $[-0.25, 0.0, 0.0]$ & Two-Stage \\
Sweep & Puck (\texttt{obj}) & Chunk 2 & $[+0.20, +0.10, 0.0]$ & Two-Stage \\
Coffee Pull & Mug \& Machine (\texttt{obj}, \texttt{coffee\_machine}) & Chunk 2 & $[-0.20, 0.0, 0.0]$ & Two-Stage \\
\bottomrule
\end{tabular}
}
\end{table}

\subsection{Details of Simulation OOD Experiments}
\label{sec:appendix_metaworld}

In this section, we provide the exact details for the Out-of-Distribution (OOD) evaluation on the MetaWorld benchmark. We detail the mechanism for implementing object displacements, how the robot tracks these shifts in simulation, and the precise task-specific parameters (timing, displacement vectors, and target objects).

\subsubsection{Oracle State Transitions and OOD Implementation}

To explicitly isolate and highlight the predictive power of the World Model's ``imagination'' capability, we simplify the simulation evaluation pipeline. Specifically, we intentionally bypass the automated $\mathcal{D}_{R-I}$ OOD detector and the heuristic recovery triggers described in the main text. Instead, we implement a controlled \textbf{Oracle Tracking and State Transition Mechanism} via direct MuJoCo backend manipulations. By hard-coding the exact timing of both the OOD anomaly injection and the physical recovery to specific action chunks, we ensure that the evaluation purely reflects the policy's ability to execute manipulation actions via internal imagination, independent of external tracking accuracy or detection latency.

During the pre-defined execution step (i.e., the specific ``chunk'' detailed in \cref{tab:sim_ood_details}), the environment execution is temporarily paused, and a sudden spatial shift is injected. As illustrated in \cref{fig:ood_disturbances_appendix} for the remaining MetaWorld tasks, the implementation follows two main logic branches depending on the nature of the task:

\begin{figure}[tb]
  \centering
  \includegraphics[height=9cm]{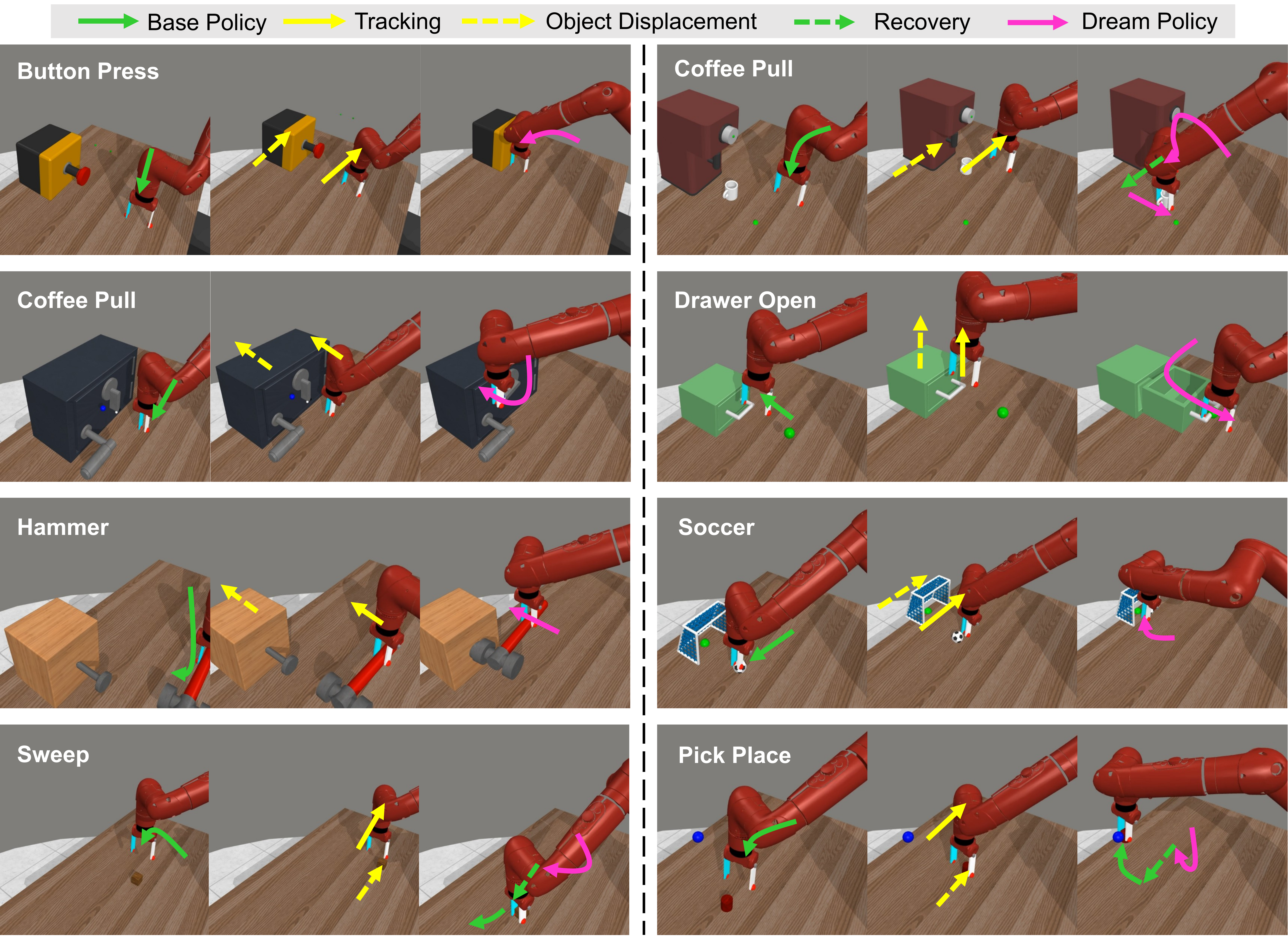}
    \caption{\textbf{DDP execution on the MetaWorld benchmark (Remaining 8 tasks).} Sequences show the transition from the ID Base Policy (green) to OOD tracking (yellow), bypassing visual shifts via the imagined Dream Policy (pink), and finally triggering physical recovery (green dotted) to complete the target.}
  \label{fig:ood_disturbances_appendix}
\end{figure}

\noindent \textbf{1. Single-Stage Tracking (Target-Centric Tasks):}
For tasks where the robot only needs to interact with the shifted object (e.g., \textit{Button Press}, \textit{Drawer Open}), we simultaneously apply a coordinate translation vector $\Delta \mathbf{P} = [dx, dy, dz]$ to both the target object's body position (\texttt{sim.model.body\_pos}) and the robot's end-effector mock-up position (\texttt{data.mocap\_pos}). By directly hard-coding this offset to the \texttt{mocap}, the robot instantly re-aligns its physical hand with the newly displaced object, maintaining the relative spatial geometry required for the internal ``imagined'' execution.

\noindent \textbf{2. Two-Stage Tracking and Recovery (Pick-and-Move Tasks):}
For multi-subtask challenges where an object is displaced but the final destination (goal) remains fixed (e.g., \textit{Pick Place}, \textit{Push}, \textit{Sweep}, \textit{Coffee Pull}), the script executes a hard-coded two-stage intervention:
\begin{itemize}
    \item \textbf{Stage 1 (Track):} At the specified anomaly chunk, the object and the robot's end-effector are shifted by $\Delta \mathbf{P}$ so the robot can successfully grasp the displaced object. (Note: For \textit{Pick Place}, a ``Safe Lift'' logic is explicitly coded—lifting the arm $5\text{cm}$, shifting it, and lowering it back—to prevent clipping through the table during teleportation).
    \item \textbf{Stage 2 (Recover and Policy Transition):} At a strictly pre-defined subsequent chunk (representing the moment the object is securely grasped), a reverse offset $-\Delta \mathbf{P}$ is applied exclusively to the robot's end-effector. This forces the robot to instantly ``snap back'' to the original ID area. Crucially, the Oracle then perfectly mimics the $\mathcal{D}_{R-I}$ re-evaluation logic defined in the main text to determine the subsequent policy mode:
    \begin{itemize}
        \item For \textit{Pick Place}, \textit{Push}, and \textit{Sweep}, bringing the grasped object back to the ID zone effectively restores global visual consistency. Consequently, the Oracle successfully re-anchors the environment and transitions the system back to the real visual stream (\textbf{ID Base Policy}).
        \item Conversely, for \textit{Coffee Pull}, the recovery step only snaps the arm and the grasped mug back to the ID target zone, while the large coffee machine is left permanently displaced. Because this leaves the global visual context strictly corrupted (which would naturally trigger $\mathcal{D}_{R-I} > \tau_{diff}$ in reality), the Oracle enforces that the system must continue executing purely via internal hallucination (\textbf{OOD Dream Policy}) to safely finish placing the mug.
    \end{itemize}
\end{itemize}

\subsubsection{Task-Specific OOD Parameters}
The exact parameters for all 10 MetaWorld tasks are summarized in \cref{tab:sim_ood_details}. The ``Intervention Timing'' refers to the specific action chunk index where the sudden displacement is injected.

\noindent \textbf{Detailed Breakdown of Multi-Subtask Interventions:}
\begin{itemize}
    \item \textbf{Pick Place:} At chunk 2, the Puck is shifted by $[-0.2, -0.1, 0]$. The robot performs a collision-free tracking maneuver. At chunk 6, the robot arm snaps back by $[+0.2, +0.1, 0]$ to place the puck into the fixed goal, \textit{successfully recovering visual consistency and resuming the ID Base Policy}.
    \item \textbf{Push:} At chunk 2, the Puck is shifted by $[-0.25, 0, 0]$. The robot tracks it. At chunk 6, the robot arm snaps back by $[+0.25, 0, 0]$ to push the object toward the original fixed target, \textit{resuming the ID Base Policy}.
    \item \textbf{Sweep:} At chunk 2, the Puck is shifted by $[+0.2, +0.1, 0]$. The robot tracks it. At chunk 6, the robot arm snaps back by $[-0.2, -0.1, 0]$ to sweep the object into the fixed hole, \textit{resuming the ID Base Policy}.
    \item \textbf{Coffee Pull:} At chunk 2, both the Mug and the Coffee Machine are shifted by $[-0.2, 0, 0]$, but the target placement zone remains fixed. At chunk 8, the robot arm snaps back by $[+0.2, 0, 0]$ to pull the mug into the correct target zone. \textit{Since the machine remains displaced (OOD), the agent continues executing via the OOD Dream Policy to complete the task.}
\end{itemize}

\subsection{Further Details of Real-World Experiments}
\label{sec:supp_real_world}

In this section, we provide additional configuration details for our real-world deployment, including the control hyper-parameters, visual processing pipeline, and the exact camera setup.

\begin{figure}[tb]
  \centering
  \includegraphics[height=6cm]{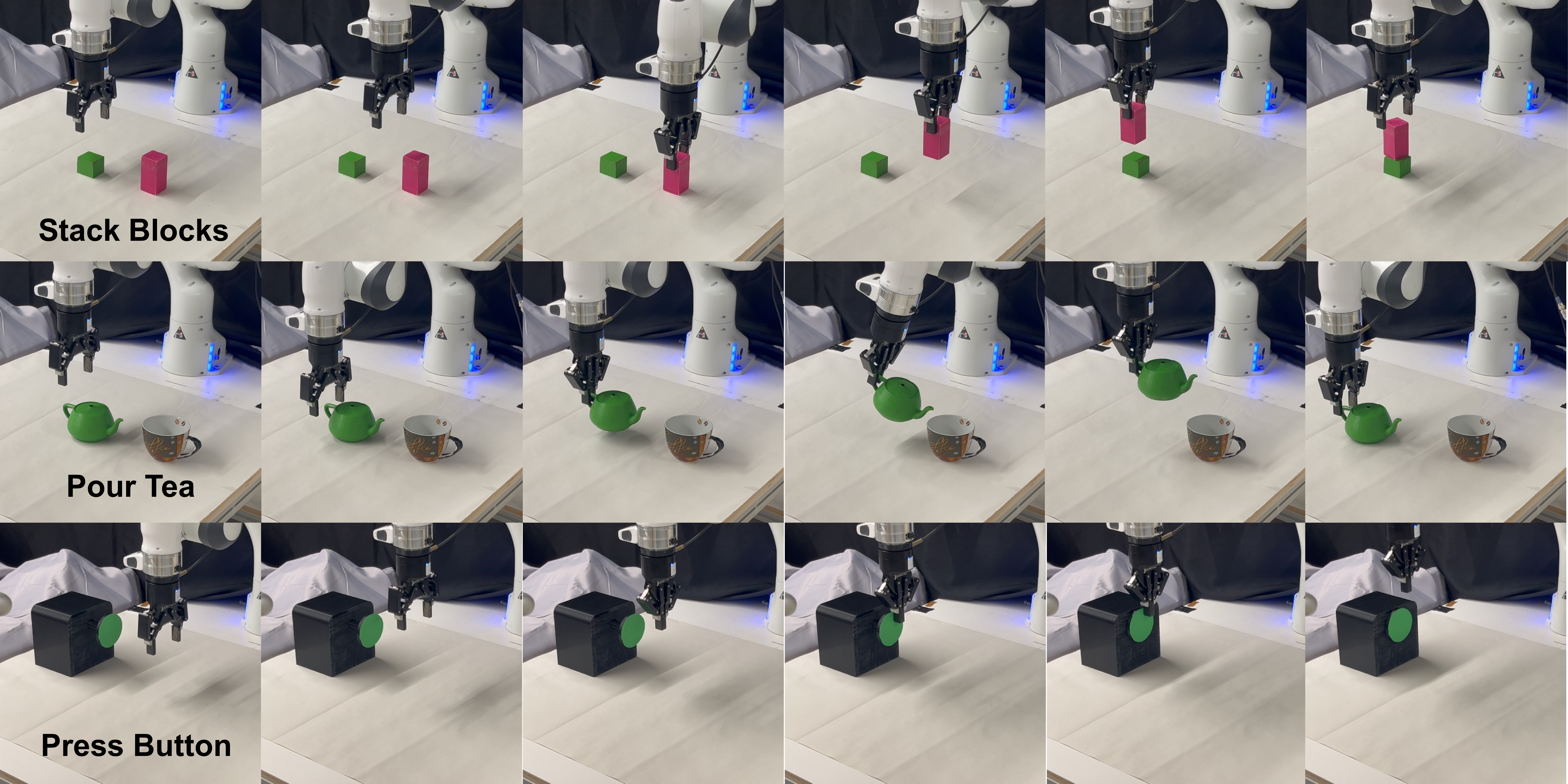}
    \caption{\textbf{In-Distribution Scenarios.} Real-world execution sequences demonstrating successful nominal manipulation for the \textit{Stack Blocks}, \textit{Pour Tea}, and \textit{Press Button} tasks.}
  \label{fig:sumplementary_id}
\end{figure}

\subsubsection{Network Inference and Control Frequency}
To achieve responsive real-time control, we accelerate the reverse diffusion process during inference by employing a fast-sampling scheduler (DDIM), reducing the denoising steps to $K=10$ for both the Diffusion Policy and the Diffusion World Model. Despite this acceleration, running the dual-diffusion pipeline alongside high-fidelity 3D visual encoding, real-time 6D object tracking, and continuous OOD anomaly detection ($\mathcal{D}_{R-I}$) is computationally intensive. Consequently, our overall system control frequency naturally bounds to $2\text{ Hz}$. Notably, this rate identically matches the frequency of our human teleoperation data collection, naturally aligning the model's temporal execution with the expert demonstrations.

We intentionally accept this inference speed to prioritize the synchronous and stable execution of all heavy subsystems. This design choice directly aligns with our primary research objective: achieving reliable, highly accurate recovery from severe OOD disturbances, rather than optimizing for high-speed dynamic manipulation. Because our evaluated tasks (e.g., \textit{Stack Blocks}, \textit{Pour Tea}) demand precise spatial reasoning and alignment rather than split-second reactive motions, the $2\text{ Hz}$ frequency proves entirely sufficient. Furthermore, the inherent temporal smoothing provided by our action chunking mechanism ($N=8$ executed steps per cycle) perfectly bridges the gap between consecutive network queries, yielding smooth, robust, and physically stable closed-loop execution.

\subsubsection{Visual Processing and Point Cloud Downsampling}
The primary visual observation is captured via a static Intel RealSense L515 camera. To achieve the required real-time processing speed, we implement a highly efficient two-stage point cloud downsampling pipeline:
\begin{enumerate}
    \item \textbf{Random Cropping:} The raw depth camera captures approximately $240,000$ points per frame. We first apply a random sub-sampling filter to drastically reduce this dense cloud to a manageable intermediate size of $10,000$ points.
    \item \textbf{Farthest Point Sampling (FPS):} We then apply FPS (accelerated via PyTorch3D) to extract a structurally uniform and geographically representative final set of exactly $2,048$ points.
\end{enumerate}
This optimized two-stage pipeline executes in just $\sim 15\text{ ms}$ on the GPU, successfully feeding high-quality geometric features into the 3D encoder without introducing bottleneck latency.

\end{document}